\PassOptionsToPackage{table}{xcolor}
\documentclass[research,10pt]{fcs}

\title{ProbDiffFlow: An Efficient Learning-Free Framework for Probabilistic Single-Image Optical Flow Estimation}
\shorttitle{Probabilistic Single-Image Optical Flow}
\author[1]{Mo ZHOU}
\author*[1]{Jianwei WANG}
\author[1]{Xuanmeng ZHANG}
\author[2]{Dylan CAMPBELL}
\author[3]{Kai WANG}
\author[4]{Long~YUAN}
\author[1]{Wenjie ZHANG}
\author*[3,1]{Xuemin LIN}
\address[1]{School of Computer Science and Engineering, University of New South Wales, Sydney 2033, Australia}
\address[2]{School of Computing, Australian National University, Canberra 2601, Australia}
\address[3]{Department of Data and Business Intelligence, Antai College of Economics \& Management, Shanghai Jiao Tong University, Shanghai 201400, China}
\address[4]{School of Computer Science and Engineering, Nanjing University of Science and Technology, Nanjing 210094, China}
\fcssetup{
  received       = {March 11, 2025},
  accepted       = {May 29, 2025},
  corr-email     = {jianwei.wang1@unsw.edu.au; xuemin.lin@sjtu.edu.cn},
}
\usepackage[ruled,vlined]{algorithm2e}
\usepackage{algorithmic}
\usepackage{adjustbox}
\usepackage{booktabs}
\SetKwInOut{KwRequire}{Require}
\SetKwInOut{KwResult}{Result}

\hypersetup{
    colorlinks=true, 
    linkcolor=blue, 
    citecolor=blue, 
    urlcolor=blue
}

\usepackage{cleveref}
\usepackage{pifont}
\usepackage{diagbox}
\usepackage{tabularx}
\usepackage{multirow}
\usepackage{colortbl}
\usepackage{placeins}
\usepackage{float}
\newcommand{\ProbDiffFlow}{\textit{ProbDiffFlow}}  
\begin{abstract}
This paper studies optical flow estimation, a critical task in motion analysis with applications in autonomous navigation, action recognition, and film production. Traditional optical flow methods require consecutive frames, which are often unavailable due to limitations in data acquisition or real-world scene disruptions. Thus, single-frame optical flow estimation is emerging in the literature. However, existing single-frame approaches suffer from two major limitations: (1) they rely on labeled training data, making them task-specific, and (2) they produce deterministic predictions, failing to capture motion uncertainty.
To overcome these challenges, we propose \ProbDiffFlow, a training-free framework that estimates optical flow distributions from a single image. Instead of directly predicting motion, \ProbDiffFlow~follows an estimation-by-synthesis paradigm: it first generates diverse plausible future frames using a diffusion-based model, then estimates motion from these synthesized samples using a pre-trained optical flow model, and finally aggregates the results into a probabilistic flow distribution. This design eliminates the need for task-specific training while capturing multiple plausible motions.
Experiments on both synthetic and real-world datasets demonstrate that \ProbDiffFlow~achieves superior accuracy, diversity, and efficiency, outperforming existing single-image and two-frame baselines.
\end{abstract}

\begin{document}
\begin{sloppypar}

\section{Introduction}

\begin{table}
\centering
\caption{Characteristics comparison of optical flow methods. }
\label{tab:Characteristics_comparison}
\resizebox{\columnwidth}{!}{
    \begin{tabular}{lcccc}
        \toprule
        \textbf{Method} & \textbf{\begin{tabular}[c]{@{}c@{}}Single-frame\\ Input?\end{tabular}} 
                        & \textbf{\begin{tabular}[c]{@{}c@{}}Probabilistic \\ Output?\end{tabular}} 
                        & \textbf{\begin{tabular}[c]{@{}c@{}}Training \\ Dependency\end{tabular}} 
                        & \textbf{\begin{tabular}[c]{@{}c@{}}Backbone\\ Architecture\end{tabular}} \\
        \midrule
        RAFT\cite{Teed2020}     & \ding{55} & \ding{55} & Requires Labeled Videos & CNN-Based \\
        PWC-Net \cite{Sun2018}         & \ding{55} & \ding{55} & Requires Labeled Videos & CNN-Based \\
        RPK-Net \cite{RPKNet2024Morimitsu}         & \ding{55} & \ding{55} & Requires Labeled Videos & CNN-Based \\
        RAPID-Flow \cite{RAPIDFlow2024}      & \ding{55} & \ding{55} & Requires Labeled Videos & CNN-Based \\
        U-Net \cite{Unet2015}           & \ding{51} & \ding{55} & Requires Labeled Images  &  CNN-Based \\
        Motion-blurred \cite{Argaw2021} & \ding{51} & \ding{55} & Requires Labeled Images & CNN-Based \\
        \midrule
        \ProbDiffFlow~(Ours)  & \ding{51} & \ding{51} & No Training Needed &  Diffusion-Based\\
        \bottomrule
    \end{tabular}
}
\end{table}

Optical flow estimation aims to predict per-pixel motion between consecutive video frames by estimating a dense displacement field~\cite{beauchemin1995computation}. This motion field represents how objects and regions move over time, providing critical information for dynamic scene understanding~\cite{wang2022reslnet}. 
{\color{blue}Optical flow has broad and significant applications in real-world scenarios; key application domains include:
\begin{itemize}[leftmargin=*, topsep=0mm]
\item \textit{Autonomous Driving}~\cite{shah2021traditional}. Accurate optical flow estimation helps detect and track moving objects in real-time, improving vehicle trajectory prediction and driving safety.
\item \textit{Surveillance Systems.} Robust optical flow estimation aids in effective crowd monitoring and timely detection of abnormal behaviors, thereby improving security measures and response efficiency.
\item \textit{Film Production and Visual Effects}~\cite{butler2012naturalistic}. Reliable optical flow methods support the accurate reconstruction of dynamic scenes, thereby enriching visual realism and significantly enhancing the overall quality of visual effects through precise scene analysis and motion tracking.
\end{itemize}
}
\noindent\textbf{Existing works}.
As summarized in Table~\ref{tab:Characteristics_comparison}, optical flow estimation falls into two categories: two-frame and single-image approaches. Two-frame methods predict motion from consecutive frames using optimization techniques~\cite{Horn1981} or deep learning models like PWC-Net and RAFT~\cite{Sun2018, Teed2020}. In contrast, single-image methods infer motion from a static frame by learning priors~\cite{2015singleflowwalker}, leveraging depth cues~\cite{Aleotti2021}, or analyzing motion blur~\cite{Argaw2021}. They all require labeled training data and produce deterministic predictions, limiting generalization and motion diversity~\cite{sun2008learning}.
\begin{figure}[t]
  \includegraphics[width=1.0\linewidth]{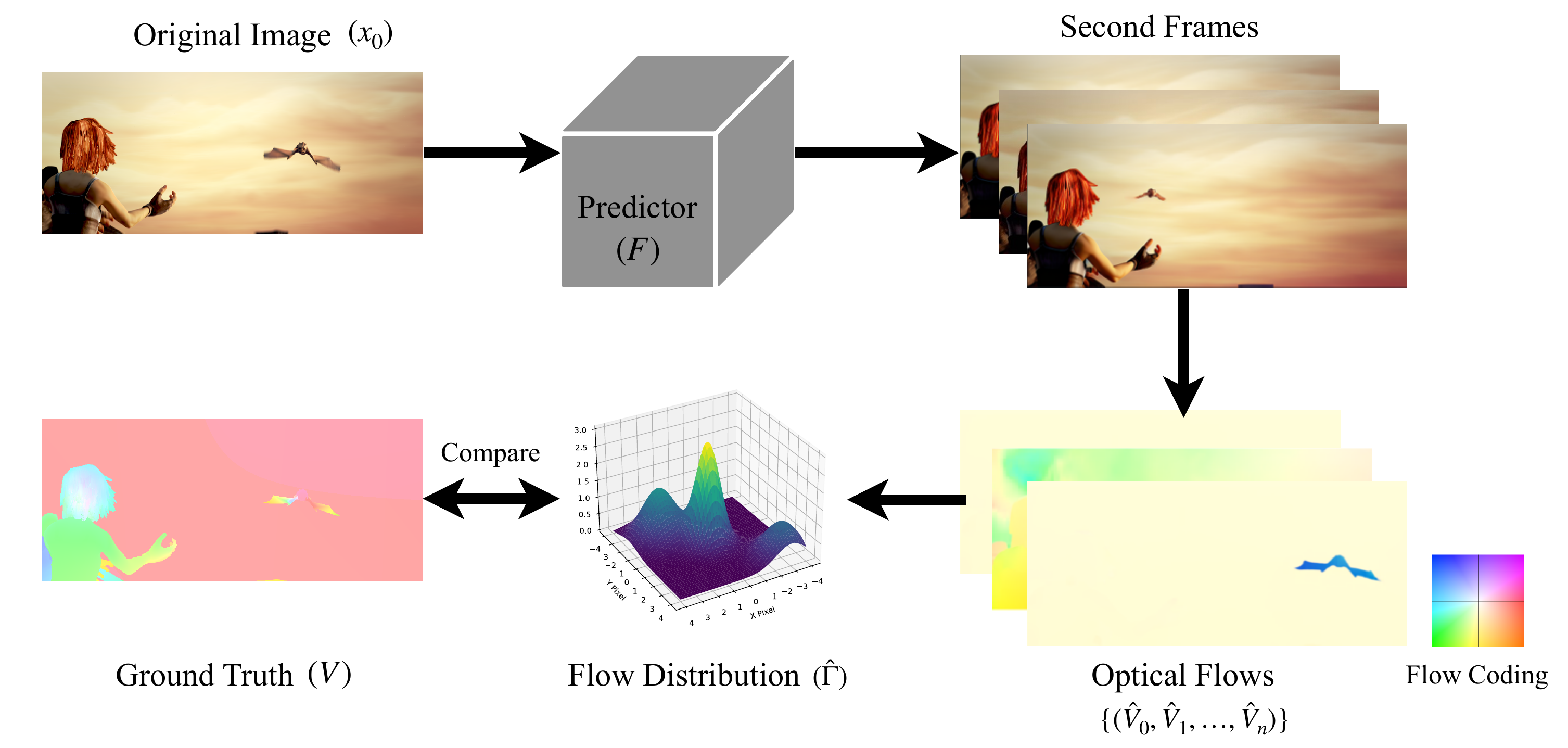}
  \caption{Problem definition. Overview of optical flow distribution prediction: Given an original image ($x_{0}$), the optical flow distribution predictor ($F$) estimates a set of plausible optical flows ($\hat{\Gamma}$).}
  \label{fig:task definition}
\end{figure}

\noindent\textbf{Motivations}.
Single-image optical flow estimation stands out for its ability to predict motion without requiring consecutive frames. It is particularly suitable for scenarios where only a single image is available, such as autonomous driving scenes with static road images and film production~\cite{denton2018stochasticvideogenerationlearned}. This flexibility expands its applicability across a wide range of domains~\cite{Wang2022RoomLayout}. However, existing single-image methods still face several critical limitations:
\begin{itemize}[leftmargin=*, topsep=0mm]
    \item \textit{Dependence on labeled training data.} Since single-frame optical flow lacks explicit motion cues, existing methods rely on supervised learning from labeled datasets to infer motion patterns~\cite{wang2020abnormal}. However, obtaining diverse, high-quality optical flow annotations is costly and labor-intensive~\cite{xu2023weakly}. As shown in~\autoref{tab:Characteristics_comparison}, current single-image approaches (\textit{e.g.,} U-Net and motion-blur-based models~\cite{Unet2015, Argaw2021}) require labeled data, making them highly task-specific and difficult to generalize across different domains~\cite{DBLP:journals/corr/abs-2501-02191,DBLP:journals/pvldb/WangWLZZ24}. 
    \item \textit{Deterministic predictions limit motion diversity.} Since existing models are trained with a single ground-truth flow per image, they produce fixed motion estimates, failing to capture the inherent uncertainty in motion prediction~\cite{ilg2018uncertaintyestimatesmultihypothesesnetworks}. However, real-world motion is often uncertain due to factors such as occlusions, complex object interactions, and depth ambiguities. These uncertainties can lead to multiple plausible motions for the same scene. A deterministic model cannot account for such variations, whereas estimating a distribution over possible motions would provide richer and more informative predictions, similar to diversity-aware approaches in other domains~\cite{Yang2019DistributedSimilarity}.
    \item \textit{Computational complexity.} 
    These single-image methods typically require training a separate model for each dataset, which is both time-consuming and computationally expensive, especially when dealing with large-scale or diverse datasets~\cite{yang2019volumetric}. Moreover, these methods often involve multiple passes over the dataset, requiring extensive optimization and hyperparameter tuning for each instance.
\end{itemize}

\noindent\textbf{Our solutions}.
To address the limitations of existing single-image optical flow methods, we propose \ProbDiffFlow, a training-free framework that estimates diverse optical flow distributions from a single image. 
The overall framework is illustrated in \autoref{fig:task definition}.
Unlike CNN-based approaches, which directly map images to fixed flow predictions, our method leverages a diffusion-based backbone to formulate optical flow as a generative problem. This eliminates the need for labeled datasets and supervised training while enabling the model to capture multiple plausible motion patterns instead of producing a single deterministic estimate. {\color{blue}In the diffusion process, \ProbDiffFlow~extends the latent-space framework with a novel nearby sampling strategy to represent motion variation across time. Specifically, we introduce a new sampler that performs unbiased perturbations in the high-dimensional latent space, enabling the generation of diverse and physically plausible future hypotheses.} This additional step empowers our model to achieve higher accuracy, greater motion diversity, and improved computational efficiency compared to previous works.

Our framework consists of three key stages: (1) \textit{Generating Plausible Subsequent Images.} Since a single static image does not contain explicit motion cues, we introduce motion diversity by synthesizing multiple possible future frames using a diffusion-based model. By defining different sampling directions and distances in the latent space, this step allows the model to explore multiple plausible motion scenarios.\newline
(2) \textit{Optical Flow Estimation from Generated Pairs.} Given the synthesized future frames, we estimate optical flow between the input and each generated frame using a pre-trained optical flow model. This step enables the framework to leverage prior motion knowledge without requiring task-specific training, making it adaptable to various motion patterns.\newline
(3) \textit{Aggregating the Optical Flow Distribution.} Instead of predicting a single deterministic flow, we analyze the frequency and consistency of motion vectors across all estimated flows to construct a per-pixel flow distribution. This provides a more comprehensive representation of possible motions, capturing motion uncertainty in dynamic scenes.

Comprehensive experiments on both synthetic and real-world datasets demonstrate that \ProbDiffFlow~outperforms both two-frame and single-frame baselines in terms of accuracy, motion diversity, and computational efficiency. It achieves lower End-Point Error (EPE), F1-score, and Angular Error (AE) while improving motion diversity. As a training-free method, it also significantly reduces computational cost and inference time compared to existing approaches, making it highly scalable for real-world applications.

\noindent\textbf{Contributions}. 
Core contributions are summarized as:
\begin{itemize}[leftmargin=*, topsep=0mm]
\item \textit{New paradigm for optical flow estimation (Section 2).} We formulate single-image optical flow distribution prediction as the task of estimating a distribution of plausible optical flows ($\hat{\Gamma}$) from a given input image $x_0$ using a flow predictor $F$. Unlike conventional methods that estimate a deterministic flow, our approach aims to approximate the unobserved conditional distribution ($\Gamma$). This requires both \textit{accuracy} (alignment with observed ground truth flow $V$) and \textit{diversity} (coverage of plausible motion hypotheses).
\item \textit{A training-free probabilistic optical flow framework (Section 3).} We propose \ProbDiffFlow, an efficient framework that predicts per-pixel motion distributions from a single image without task-specific training. It synthesizes diverse subsequent frames using a diffusion-based sampling method and computes optical flow results. 
\item \textit{Comprehensive empirical validation (Section 4).} {\color{blue}We evaluate \ProbDiffFlow~using standard metrics on multiple synthetic and real-world benchmarks, confirming its effectiveness and broad applicability.}
\end{itemize}
Our code is available at:\newline
\url{https://anonymous.4open.science/r/Single-frame-optical-flow-with-IMPUS-FB07/}.

\section{Preliminaries}

This section formalizes the problem of optical flow distribution prediction and establishes the theoretical foundation by reviewing three key diffusion models.

\subsection{Problem Statement}

\begin{table}[t]
    \centering
    \caption{Symbols and Descriptions}
    \label{tab:notation}
    \begin{tabular}{cl}
        \toprule
        \textbf{Notation} & \textbf{Description} \\
        \midrule
        \(x_0\)                      & original image (\(1^{\text{st}}\) frame)\\
        \(x_{i}\)                    & \(2^{\text{nd}}\) frames \\
        \(f_{\phi}\)                 & text encoder   \\
        \(\mathcal{L}_{\text{DPM}}\) & pre-trained Diffusion Probabilistic Model \\
        \(e^{0}\)                    & text embedding \\
        \(z_{0}\)                    & latent state of the \(1^{\text{st}}\) frame \\    
        \(z_{i}^{(t)}\)                    & latent state of the \(i^{\text{th}}\) frames at timestep $t$\\
        \(\alpha_1\)                 & learning rate \\
        \(\eta\)                     & perturbation direction vector in the latent space \\
        \(T\)                        & backward diffusion steps \\
        \(T_{\text{inv}}\)           & inversion steps in forward diffusion process \\
        \(F\)                        & pre-trained optical flow predictor \\
        \(\hat{V_i}\)                & predicted \(i^{\text{th}}\) flow \\
        \(V\)                        & ground truth flow \\
        \(\hat{\Gamma}\)             & predicted flow distribution \\
        \(\Gamma\)                   & unobserved conditional flow distribution \\
        \bottomrule
    \end{tabular}
\end{table}

Given a single input image $x_0$, our objective is to predict a distribution of plausible optical flows, denoted as $\hat{\Gamma}$ by using flow predictor $F$ as illustrated in~\autoref{fig:task definition}. This can be defined mathematically as: 
\begin{equation}
    \hat{\Gamma} = F(x_0) = \{\hat{V_0}, \hat{V_1}, \dots, \hat{V_n}\},
\end{equation} 
where $\hat{V_i}$ denotes the $i^{\text{th}}$ predicted optical flow hypothesis and $V$ represents the single ground truth flow. The predicted distribution $\hat{\Gamma}$ should approximate the unobserved conditional distribution $\Gamma$ that encapsulates all physically valid flow solutions, requiring both \textit{accuracy} (alignment with observed $V$) and \textit{diversity} (coverage of plausible results). Complete symbol definitions are provided in~\autoref{tab:notation}.

\subsection{Diffusion Models}

Modern diffusion models derive from two seminal works: denoising diffusion probabilistic models (DDPMs)~\cite{ho2020denoising} and their accelerated variant, denoising diffusion implicit models (DDIMs)~\cite{song2022denoising}. Later extensions introduce the latent space formulation of stable diffusion~\cite{stablediffusion}, which inspires the architectural design of this work.

\noindent \textbf{DDPM \& DDIM Foundations.}
The DDPM framework establishes two core processes:
\begin{itemize}[leftmargin=*, topsep=0mm]
    \item \textit{Forward diffusion}: Gradually converts data \(x^{(0)}\) through \(T\) noise-adding steps:
    \begin{equation}
        x^{(t)} = \sqrt{\alpha^{(t)}} x^{(t-1)} + \sqrt{1-\alpha^{(t)}} \epsilon, \quad \epsilon \sim \mathcal{N} (0,1).
    \end{equation}
    A more direct calculation method is given by:
    \begin{equation}
        x^{(t)}=\sqrt{\bar{\alpha}^{(t)}} x^{(0)}+\sqrt{1-\bar{\alpha}^{(t)}} \epsilon,
    \end{equation}
    where \(\bar{\alpha}^{(t)}=\prod_{i=1}^t\left(\alpha^{(i)}\right) \) represents the cumulative noise schedule.
    \item \textit{Reverse process}: Learns to iteratively denoise through:
    \begin{equation}
        p_\theta\left(x^{(t-1)} \mid x^{(t)}\right)=\mathcal{N}\left(x^{(t-1)} ; \mu_\theta\left(x^{(t)}, t\right), \Sigma_\theta\left(x^{(t)}, t\right)\right).
    \end{equation}
\end{itemize}

The Denoising Diffusion Probabilistic Model (DDPM) relies on a Markov chain, which slows down the inference process due to the high number of iterations. To address this limitation, the Denoising Diffusion Implicit Model (DDIM) introduces a method that accelerates the sampling process by breaking the Markov chain. The forward process in DDIM remains identical to that of DDPM, while the reverse process is formulated as follows:
\begin{equation}
    \begin{split}
        x^{(t-1)} = &\sqrt{\bar{\alpha}^{(t-1)}} \frac{x^{(t)}-\sqrt{1-\bar{\alpha}^{(t)}} \hat{\epsilon}_\theta\left(x^{(t)}, t\right)}{\sqrt{\bar{\alpha}^{(t)}}} \\
        &+\sqrt{1-\bar{\alpha}^{(t-1)}} \hat{\epsilon}_\theta\left(x^{(t)}, t\right).
    \end{split}
\end{equation}
\[  \]
Here, we sample timesteps \(t \in \{1,2,\dots T_{\text{inv}}\}\).

\noindent \textbf{Stable Diffusion's Latent Space Innovation.}
%
Stable Diffusion generates images by performing the diffusion process in a latent space~\cite{stablediffusion}. It first encodes an image ($x^{(0)}$) into a latent representation ($z^{(0)}$) using a variational autoencoder (VAE), where noise is iteratively added and removed during the forward and reverse processes. The reverse process is conditioned on a text embedding ($e^{(0)}$), obtained using a pre-trained contrastive language-image pre-training (CLIP) model, while a U-Net backbone refines the latent variables by predicting and removing noise. The iterative denoising process follows:
\begin{equation}
\begin{split}
    z^{(t-1)}  =& \sqrt{\bar{\alpha}^{(t-1)}} \frac{z^{(t)}-\sqrt{1-\bar{\alpha}^{(t)}} \hat{\epsilon}_\theta\left(z^{(t)}, e^{(0)}, t\right)}{\sqrt{\bar{\alpha}^{(t)}}} \\
    &+\sqrt{1-\bar{\alpha}^{(t-1)}} \hat{\epsilon}_\theta\left(z^{(t)}, e^{(0)}, t\right).
\end{split}
\end{equation}
This formulation enables image synthesis within a latent space, supporting our optical flow prediction framework.

{\color{blue}
While our framework builds upon Stable Diffusion, it introduces a key innovation to adapt diffusion models for motion estimation: a novel nearby sampling strategy in the latent space. We perturb the latent state of the original image in controlled directions and distances. This process enables us to simulate plausible temporal changes from a single frame and construct diverse future hypotheses, making the model suitable for probabilistic single-image optical flow estimation.}

\section{Method}

\begin{figure*}[t]
  \centering
  \includegraphics[width=\linewidth]{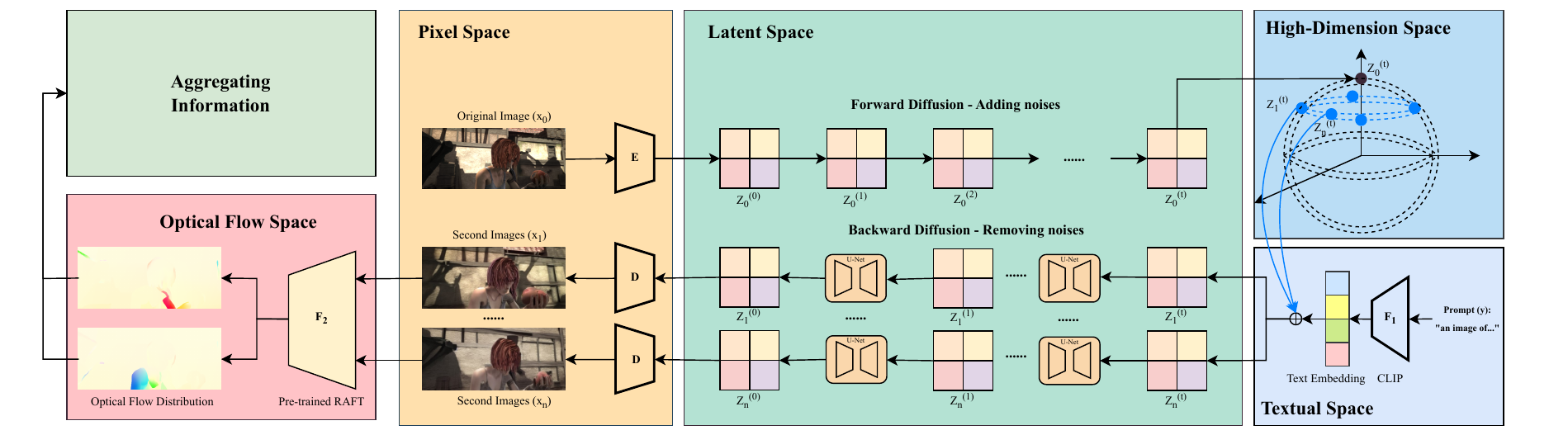}
  \caption{Overall framework of \ProbDiffFlow. Given a single image \(x_0\), the method proceeds as follows: (a) The image is encoded into a latent representation \(z_0^{(0)}\) using a VAE, and a textual embedding \(e^{(0)}\) is obtained for conditioning; (b) Forward diffusion adds noise to produce the noisy latent state \(z_0^{(t)}\); (c) A nearby sampling strategy perturbs \(z_0^{(t)}\) to generate a set of diverse latent vectors \(\{z_i^{(t)}\}\) on the same high-dimensional spherical shell; (d) Each latent is denoised through reverse diffusion to generate future frame candidates \(\{x_i\}\); (e) Optical flow between \(x_0\) and each \(x_i\) is computed via a pre-trained RAFT model, and the results are aggregated into a probabilistic flow distribution \(\hat{\Gamma}\).}
  \label{fig:overall framework}
\end{figure*}

In this section, we introduce \ProbDiffFlow, a training-free method for predicting an optical flow distribution from a single image. The key idea behind our approach is using the power of a diffusion model to generate multiple plausible outcomes without requiring extensive retraining. Our framework consists of three main processes: generating plausible nearby images, optical flow estimation from generated pairs, and aggregating the optical flow distribution.

Unlike other methods that rely on CNN-based backbones and require training from scratch with large labeled datasets, substantial computational resources, and long training times, our approach is training-free, eliminating the need for model optimization. Instead, it leverages a diffusion-based backbone, utilizing the generative capabilities of Stable Diffusion for image synthesis \cite{stablediffusion}. This reduces computational cost and complexity while enabling our model to generate diverse optical flow distributions rather than a single deterministic prediction. By capturing multiple plausible motion hypotheses, our method enhances efficiency and flexibility, making it well-suited for real-world scenarios.

{\color{blue}\subsection{Overall Framework and Algorithm}

The overall framework is illustrated in~\autoref{fig:overall framework} and the process is summarized in~\autoref{alg:probFlowDiff}. It takes a single input image $x_0$ and outputs an estimated optical flow distribution $\hat{\Gamma}$. The algorithm follows a structured pipeline to efficiently generate plausible motion hypotheses without requiring any training. The steps are as follows:
\begin{itemize}[leftmargin=10pt, topsep=1pt]
\item[a.] \textit{Image Encoding (Phase 1; Fig. 2(a)):} The original image $x_{0}$ is encoded into a latent representation $z_{0}^{(0)}$ using a variational autoencoder (VAE). Meanwhile, a text embedding $e^{(0)}$ is obtained or refined using textual inversion to guide the diffusion process (\textit{lines 1–2}).
\item[b.] \textit{Forward Diffusion (Phase 2; Fig. 2(b)):} A forward diffusion process adds noise to the latent vector through $T_{inv}$ timesteps, resulting in a noisy latent state $z_0^{(t)}$ \textit{(lines 3–8)}.
\item[c.] \textit{Latent Nearby Sampling (Phase 3; Fig. 2(c)):} We perturb $z_0^{(t)}$ along multiple directions with fixed magnitude using our nearby sampling strategy. This generates a set of perturbed latent states $\{z_{i}^{t}\}$ that remain on the same high-dimensional spherical shell, ensuring unbiased and realistic variation \textit{(lines 9–13)}.
\item[d.] \textit{Backward Diffusion and Image Decoding (Phase 4 and Phase 5; Fig. 2(d)):} Each sampled latent state is passed through the reverse diffusion process, conditioned on the shared embedding $e^{(0)}$, to generate diverse future images $\{x_{i}\}$ \textit{(lines 15–20)}.
\item[e.] \textit{Flow Estimation and Aggregation (Phase 6 and Phase 7; Fig. 2(e)):} Each generated frame $x_{i}$ is paired with the original $x_{0}$, and a pre-trained RAFT model is used to compute flow predictions $\{\hat{V_{i}}\}$ \textit{(lines 21–22)}. Instead of selecting one deterministic result, we aggregate all estimated flows into a probabilistic distribution $\hat{\Gamma}$ to capture motion uncertainty \textit{(line 23)}. Summary statistics and diversity metrics are computed in the final step \textit{(lines 24–25)}.
\end{itemize}}


\begin{algorithm}[htbp]
\SetAlgoLined    
\LinesNumbered    
\caption{\ProbDiffFlow: An Efficient Learning-Free Framework for Probabilistic Single-Image Optical Flow Estimation}
\label{alg:probFlowDiff}
\resizebox{\linewidth}{!}{ 
\begin{minipage}{\linewidth}
\SetKwInOut{Input}{Input}
\SetKwInOut{Output}{Output}

\Input{
Input original image ($x_0$)
}
\Output{Estimated optical flow distribution ($\hat{\Gamma}$)}

{\color{blue}\tcc{Initialize}
$T_{inv} = 250$;
$N = 500$\;}

\tcc{Phase 1: Image Encoding to Latent Space}
$z_0^{(0)} \leftarrow \mathrm{VAE.encode}(x_0)$\;
$e^{(0)} \leftarrow f_\phi(y)$\;

\tcc{Phase 2: Textual Inversion and Forward Diffusion}
\For{$i \gets 1$ \KwTo $T_{\text{inv}}$}{
  Sample time step $t \sim \{1, \dots, T_{\text{inv}}\}$\;
  Sample random noise $\epsilon_t \sim \mathcal{N}(0, I)$\;
  $z_0^{(t)} \leftarrow \sqrt{\beta_t}\,z_0^{(0)} + \sqrt{1 - \beta_t}\,\epsilon_t$\;
  $e^{(0)} \leftarrow e^{(0)} - \alpha_1\,\nabla_{e^{(0)}}\,\mathcal{L}_{\mathrm{DPM}}(z_0^{(0)}, e^{(0)};\theta)$\;
}

\tcc{Phase 3: Latent Sampling in Nearby Space}
Compute direction vector $d$ \;
\For{$i \gets 1$ \KwTo $N$}{
  $z_i^{(t)} \leftarrow z_0^{(t)} + \delta\,d$\;
  Reproject $z_i^{(t)}$ to same spherical norm as $z_0^{(t)}$\;
}

\tcc{Phase 4: Backward Diffusion Process}
\For{$i \gets 1$ \KwTo $N$}{
  \For{$t \gets T$ \KwTo $1$}{
    $z_{i}^{(t-1)} \leftarrow \sqrt{\beta_{t-1}}
      \Bigl(\dfrac{z_i^{(t)} - \sqrt{1-\beta_t}\,\hat{\epsilon}_\theta^{(t)}(z_i^{(t)}, e^{(0)})}{\sqrt{\beta_t}}\Bigr)
      \;+\;\sqrt{1-\beta_{t-1}}\,\hat{\epsilon}_\theta^{(t)}(z_i^{(t)}, e^{(0)})$\;
  }
  {\color{blue}\tcc{Phase 5: Decoding Latent to Image Space}
  $x_i \leftarrow \mathrm{VAE.decode}(z_i^{(0)})$\;}
}

\tcc{Phase 6: Flow Distribution Estimation}
Construct pairs $(x_0, x_i)$ for $i=1,\dots,N$\;
$\hat{V}_i \leftarrow F_{\mathrm{RAFT}}(x_0, x_i)$\;
$\hat{\Gamma} = \{\hat{V}_i \mid i=1,\dots,N\}$\;

\tcc{Phase 7: Information Aggregation}
Compute \textit{accuracy (EPE, F1-all)} and \textit{diversity (entropy)} of $\hat{\Gamma}$\;

\Return $\hat{\Gamma}$\;
\end{minipage}
}
\end{algorithm}

\subsection{Generating Plausible Nearby Images}

We employ Stable Diffusion to generate plausible nearby images. This process consists of four main steps: (1) textual inversion, (2) forward diffusion, (3) nearby sampling, and (4) reverse diffusion. Specifically, we use the DDIM scheduler for both forward and reverse diffusion steps, enabling an efficient generation of realistic variations. Each step contributes to generating a diverse set of images that are plausible variations of the original, capturing the potential outcomes in the latent space effectively.

\noindent \textbf{Textual Inversion.}
In the stable diffusion model, text inversion involves embedding the text and optimizing this embedding to guide the diffusion process. Given an image $x_0$, we manually define a common prompt $y$: $<\text{a photo of }[\text{token}]>$, where a token is the object class. It is used to generate a text embedding that semantically represents the image. Using the pre-trained CLIP model \cite{clip2021}, the text is embedded into the same latent space as the image, expressed as $e^{(0)} \leftarrow f_\phi(y)$.

The objective is to find the optimal embedding $e^{(0)}$ that accurately represents the original image during the diffusion process with the pre-trained diffusion probabilistic model (DPM). The optimization formula is:
\begin{equation}
    e^{(0)} = \arg \min_{e} \mathcal{L}_{\text{DPM}} (z_0^{(0)}, e; \theta),
\end{equation}
where $\mathcal{L}_{\text{DPM}}$ refers to the DPM loss function, $z_0^{(0)}$ refers to the latent state of the original image and $\theta$ represents the model parameters. During the backward process, the text embedding $e^{(0)}$ conditions the noise prediction at timestep $t$, guiding the image generation to align with the text description. {\color{blue}This semantic guidance ensures that the generated images maintain high-level consistency with the input image class and scene, preventing semantic distortion during generation.}

\noindent \textbf{Forward Process.}
Unlike previous diffusion models that add noise directly in pixel space, we follow the stable diffusion approach by encoding the original image into a latent space. Specifically, we use a variational autoencoder \(E\) to encode the original image \(x_0\) into its latent representation $z_0^{(0)}$. This step can compress the image without losing information. Noise is then added to the latent space, making the forward process more efficient and effective for capturing high-dimensional structures. The forward process can be expressed as:
\begin{equation}
    z_0^{(t)} = \sqrt{\beta_t} z_0^{(0)} + \sqrt{1 - \beta_t} \epsilon,
\end{equation}
where \( \epsilon \sim \mathcal{N}(0, I) \) is a gaussian noise, and \( \beta_t \) represents the cumulative noise schedule. This can be parameterized as:
\begin{equation}
    z_0^{(t)} \sim q(z_0^{(t)} | z_0^{(0)}) = \mathcal{N}(\sqrt{\beta_t} z_0^{(0)}, (1 - \beta_t) I).
\end{equation}

This parameterization provides a probabilistic perspective of how noise is gradually added to the latent embedding, making the model capable of handling a wide range of complex image distributions. 

\noindent \textbf{Nearby Sampling.}
Inspired by the interpolating latent space process, we propose a novel nearby sampling method in the latent space. At the final step of the forward process (timestep $t$), we obtain the latent representation of the original image, denoted as $z_0^{(t)}$. The next step involves sampling multiple nearby latent states $z_i^{(t)}$, where $i$ represents the $i$-th sampled point. These samples are generated by perturbing $z_0^{(t)}$ within a fixed sampling distance while varying the sampling direction, ensuring diverse yet controlled variations in the latent space. Since $z_0^{(t)}$ is obtained through the forward diffusion process, it follows a high-dimensional Gaussian distribution, where most of the probability mass is concentrated in a thin spherical shell. This property motivates us to reproject the sampled states, ensuring they remain on the same spherical shell and follow the same distribution as the original state. To achieve this, we define the following steps for generating new sample points in the latent space:

\begin{enumerate}[leftmargin=0pt,itemindent=2em,label=\arabic*)]
    \item \textit{Generate a Random Perturbation Direction Vector ($\eta$).} We begin by sampling a random vector $\eta \sim \mathcal{N}(0,1)$, {\color{blue}drawn from a standard multivariate normal distribution. This Gaussian distribution is isotropic and centered at the origin, meaning that the direction of $\eta$ is uniformly distributed over the high-dimensional unit sphere. As a result, the sampling is unbiased and covers all possible directions uniformly.}
    \item \textit{Project \(\eta\) onto the Tangent Plane.}  Since we are working in a spherical latent space, a direct perturbation using \(\eta\) may push the sampled states off the desired manifold. To address this, we project \(\eta\) onto the tangent plane of the original state \(z_0^{(t)}\). The projection ensures that the perturbation is orthogonal to the radial component of \(z_0^{(t)}\), thereby maintaining consistency with the latent space structure. This is achieved using the formula:
    \begin{equation}\label{Eq:4.2}
        \eta \leftarrow \left(I - \frac{z_0^{(t)}{z_0^{(t)}}^{T}}{ {\parallel z_0^ {(t)} \parallel} ^{2}}\right) \eta,
    \end{equation}
    here, \(I\) is the identity matrix.
    \item \textit{Get Unit Direction Vector (\(d\)).}  The perturbation vector \(\eta\) is normalized to obtain a unit direction vector \(d\). This normalization ensures that all sampled perturbations have uniform step sizes, preventing distortions due to varying magnitudes of \(\eta\). The unit vector is computed as: \( d = \frac{\eta}{\parallel \eta \parallel}, \) where \(\parallel \eta \parallel\) denotes the Euclidean norm of the perturbation vector.
    \item \textit{Calculate the Latent State (\(z_i^{(t)}\)):}  To generate multiple nearby states, we introduce a constant perturbation distance \(\delta\), ensuring that all sampled states are equidistant from \(z_0^{(t)}\). Each new latent state \(z_i^{(t)}\) is computed as: 
    \begin{equation}
        z_i^{(t)} \leftarrow z_0^{(t)} + \delta d,
    \end{equation}
    where \(i \in \{1, 2, 3, \dots, n\}\) is the index of the generated states, and \(n\) represents the total number of sampled points. The parameter \(\delta\) controls the distance between each sampled state and the original point, which directly influences the degree of deformation in the newly generated images.
    \item \textit{Reproject to the Sphere.}  Since the previous step moves the sampled points slightly away from the original spherical manifold, we perform a final re-projection to ensure that \(z_i^{(t)}\) remains on the same spherical shell as \(z_0^{(t)}\). This is done by normalizing \(z_i^{(t)}\) and rescaling it to the norm of \(z_0^{(t)}\), using:
    \begin{equation}
        z_i^{(t)} \leftarrow z_i^{(t)} \frac{\parallel z_0^{(t)} \parallel}{\parallel z_i^{(t)}\parallel}.
    \end{equation}
    This correction guarantees that the sampled latent states adhere to the original latent distribution, preserving consistency across different perturbation directions.
\end{enumerate}
{\color{blue}To summarize, semantic similarity is achieved by using the same textual prompt and textual embedding, which places the generated image near the original image in the latent space. Nearby sampling further controls structural diversity by introducing isotropic perturbations around this position. Together, these mechanisms allow \textit{ProbDiffFlow} to generate second images that are both semantically aligned and structurally consistent with the original.}

\noindent \textbf{Reverse Process.}
The reverse diffusion process removes noise from the latent data to reconstruct an image that closely reaches the original image. In the DDIM reverse scheduler, the update rule for each timestep is derived from the conditional distribution: \( q(z_0^{(t)} \mid z_0^{(0)}) := \mathcal{N} (z_0^{(t)}; \beta_t z_0^{(0)}, (1-\beta_t) I) \), and can be expressed as follows:
\begin{equation}
\begin{split}
    z_{i}^{(t-1)} = &\ \sqrt{\alpha_{t-1}} \left(\frac{z_i^{(t)} - \sqrt{1 - \alpha_t} \hat{\epsilon}_\theta^{(t)}(z_i^{(t)}, e^{(0)})}{\sqrt{\alpha_t}}\right) \\
    &+ \sqrt{1 - \alpha_{t-1}} \hat{\epsilon}_\theta^{(t)}(z_i^{(t)}, e^{(0)}),
\end{split}
\end{equation}
where \( \hat{\epsilon}_\theta^{(t)} \) represents the parameterized noise predictor, and the text embedding \( e^{(0)} \) serves as a conditioning factor alongside the latent state from the previous step (\( z_i^{(t)} \)). The goal of the reverse diffusion process is to minimize the discrepancy between the predicted noise and the actual noise added during the forward process. The objective function for DPM, which aligns with the forward process, is given by:
\begin{small}
\begin{equation}
   \min_{\theta}\mathcal{L}_{\text{DPM}}(z_0^{(0)}, e^{(0)}; \theta) = \min_{\theta} \mathbb{E}_{z_0^{(t)}, \epsilon, t} \left[ \left\| \epsilon - \hat{\epsilon}_\theta(z_0^{(t)}, e^{(0)}, t) \right\|^2_2 \right].
\end{equation}
\end{small}
After obtaining the denoised latent embedding (\( z_i^{(0)} \)), it is decoded back into pixel space (\(x_i\)) using the VAE decoder.

\subsection{Optical Flow Estimation from Generated Pairs}
Due to the lack of a reference frame in single-image optical flow estimation, we generate multiple plausible subsequent images to approximate different motion hypotheses. Each generated image represents a possible future frame, enabling a distribution of flow predictions rather than a single deterministic output.

Given the original image \(x_0\) and the generated subsequent images \( \{ x_i \mid i \in \{ 1, 2, 3, \dots, N \} \} \), we create $N$ image pairs by pairing each generated image with the original, resulting in: $((x_0, x_1), (x_0, x_2), \dots, (x_0, x_N))$, each pair represents the potential motion from the original image to the generated one.

To estimate the optical flow for each pair, we utilize a pre-trained recurrent all-pairs field transform (RAFT), a well-established model in the optical flow domain~\cite{Teed2020}. RAFT employs iterative refinement and an all-pairs correlation mechanism, enabling robust motion estimation even in the presence of occlusions and large displacements. For each image pair \( (x_0, x_i) \), RAFT directly predicts a dense optical flow map \( \hat{V_i} \) that describes the pixel-wise displacement between the images. This process can be summarized as:
\begin{equation}
    \hat{V_i} = F_{\text{RAFT}} (x_0, x_i), \quad i \in \{1, 2, 3, \dots, N\}.
\end{equation}
Instead of selecting a single best flow estimate, we retain all $N$ predictions to model the full distribution of plausible motion. This serves as the foundation for the subsequent aggregation step, where we extract meaningful motion patterns from the generated flow samples.\\

\subsection{Aggregating the Optical Flow Distribution}
Real-world motions are inherently uncertain, particularly in dynamic scenes with moving objects. A single deterministic flow prediction cannot capture the full range of possible motions. For example, a vehicle at an intersection may move in multiple directions~\cite{chai2019multipath}, and occlusions can introduce further ambiguity~\cite{ilg2018occlusions}. To better model motion uncertainty, we aggregate multiple optical flow predictions into a probabilistic distribution, providing a more comprehensive representation of potential movement patterns.

To analyze this distribution, we visualize the predicted optical flow directions and magnitudes using polar histograms. In this representation, the length of each bar indicates the number of pixels moving in a specific direction, highlighting dominant motion trends. The color encoding represents the mean magnitude of motion in each direction, revealing variations in motion intensity across different regions.

By examining the aggregated flow distribution, we gain a clearer understanding of motion uncertainty and dominant movement patterns. Additionally, we evaluate the optical flow distribution using entropy to measure motion diversity and standard accuracy metrics such as EPE and F1-all. These results, presented in~\autoref{tab:comparison} and~\autoref{tab:entropy}, demonstrate that our method effectively captures diverse motion hypotheses while maintaining strong prediction accuracy.

\section{Experiment}

\subsection{Datasets Description}
Our experiments leverage three established benchmarks representing distinct optical flow challenges:
\begin{itemize}[leftmargin=*, topsep=0mm]
    \item MPI Sintel~\cite{Butler:ECCV:2012} is derived from the animated film Sintel and widely used in optical flow research due to its challenging and diverse sequences. We use 23 sequences with 1042 images, covering complex motion, large displacements, and realistic motion blur. It is divided into Sintel Clean, which focuses on smooth and clear motion, and Sintel Final, which introduces additional motion complexity.
    \item KITTI 2015~\cite{Geiger2012CVPR} consists of 200 real-world urban driving scenes (1242×375 resolution), evaluating optical flow performance in dynamic environments. The dataset is widely used for assessing flow estimation in autonomous driving scenarios, featuring real-world occlusions, moving objects, and large depth variations.
    \item Spring~\cite{spring2023} is a high-resolution (1920×1080) dataset designed for fine-grained optical flow estimation. It presents complex, photo-realistic motion patterns, making it a valuable benchmark for evaluating the ability of models to capture subtle motion details.
\end{itemize}
These datasets provide high-quality ground truth optical flow data and diverse scene compositions, offering an ideal setting to assess the capability of our method in predicting realistic motion from a single image.

\subsection{Experimental Setup}
\noindent \textbf{Baselines.}
We compare our method with representative baselines categorized into two types: traditional two-frame CNN-based methods and single-frame methods.
\begin{itemize}[leftmargin=*, topsep=0mm]
    \item \textit{Two-frame CNN-based Methods:} These models require consecutive image pairs $(x_0, x_1)$ as input and estimate optical flow between them. We consider RAFT~\cite{Teed2020}, PWC-Net~\cite{Sun2018}, RPK-Net~\cite{RPKNet2024Morimitsu}, and RAPID-Flow~\cite{RAPIDFlow2024}. To enable a fair comparison with our single-frame approach, we adapt these models by using stable video diffusion (SVD) to generate plausible subsequent frames $\hat{x_1}$~\cite{svd2023Andreas}. This allows us to synthesize frame pairs $(x_0, \hat{x_1})$, aligning the input format of conventional methods with our single-frame setting while preserving their original architectures.
    \item \textit{Single-frame Methods:} To further benchmark our method, we implement a single-frame baseline that estimates optical flow from a motion-blurred image using an end-to-end encoder-decoder architecture~\cite{Argaw2021}. U-Net, originally developed for image segmentation, serves as a widely used backbone for pixel-level predictions~\cite{Unet2015}. We adapt it for single-image optical flow estimation, providing a strong baseline to assess the effectiveness of our approach.
\end{itemize}

\noindent \textbf{Metrics.}
Our evaluation considers three key aspects: accuracy, diversity, and efficiency. For accuracy, we use End-Point Error (EPE), F1-all, and Angular Error (AE), where lower values indicate better performance. These metrics complement each other by capturing different sensitivity patterns in motion error evaluation~\cite{Fortun2015}. For diversity, higher entropy indicates richer variation in predicted motion. Efficiency is measured by runtime and FLOPs, with lower values reflecting better computational performance. Below, we detail the specific evaluation metrics used in our study.
\begin{itemize}[leftmargin=*, topsep=0mm]
    \item \textit{Accuracy.} We assess optical flow estimation performance using the following metrics:
        \begin{itemize}[leftmargin=*, topsep=0mm]
            \item \textit{End-Point Error (EPE, px).} EPE is a standard metric for optical flow evaluation. It measures the Euclidean distance between the estimated flow vector $v_p$ and the ground truth flow vector $v_{gt}$. For a single prediction, EPE is computed as:
            \begin{equation}
                EPE = \sqrt{(v_{gt, 0}-v_{p, 0})^2 + ( v_{gt, 1}-v_{p, 1})^2}.
            \end{equation}
            Since our method generates $N$ flow predictions per pixel $(i,j)$, we compute the average EPE across all generated predictions:
            \begin{equation}
                EPE_{avg} = \frac{1}{N}\sum_{k=1}^{N}\sqrt{(v_{gt, 0}-v_{p_k, 0})^2 + ( v_{gt, 1}-v_{p_k, 1})^2}.
            \end{equation}
            A lower EPE indicates a closer match between the predicted and actual flow.
            \item \textit{F1-all (\%).} F1-all measures the percentage of pixels with large flow errors. A predicted flow is classified as an outlier if its EPE exceeds 3 pixels and is also greater than 5\% of the ground truth motion magnitude. In addition to F1-all, we report F1-fg (foreground outliers) and F1-bg (background outliers) to analyze performance across different motion regions.
            \item \textit{Angular Error (AE, $^\circ$).} AE quantifies the angular difference between the predicted and ground truth flow vectors, measuring directional similarity. It is computed using cosine similarity:
            \begin{equation}
                AE = \cos^{-1}(\frac{v_p^Tv_{\text{gt}}}{||v_p||||v_{\text{gt}}||}).
            \end{equation}
            The AE ranges from $0^\circ$ to $180^\circ$, where $0^\circ$ indicates perfect alignment and $180^\circ$ represents opposite directions.
        \end{itemize}
    \item \textit{Diversity (Entropy).} To assess the ability of models to generate diverse motion predictions, we compute entropy over the predicted optical flow distribution. For a given pixel $(i,j)$, our method generates $N$ flow vectors in a 2D space. We discretize this space into a grid of size $h \times w$, where each cell represents a motion direction and magnitude bin. The probability of a flow vector falling into a specific bin $(i,j)$ is given by: $p_{i,j} = \frac{n_{i,j}}{n}$, where $n_{i,j}$ is the number of flow vectors in that bin. The entropy is then computed as:
    \begin{equation}
        H(V) = - \frac{1}{\log (h \times w)} \sum_{i=1}^h \sum_{j=1}^w p(c_{i,j}) \log p(c_{i,j}).
    \end{equation}
    The normalization factor $\log(h \times w)$ ensures that the entropy value lies within the range $[0, 1]$, where 0 represents a fully deterministic prediction and 1 indicates maximally diverse flow estimates.
    \item \textit{Efficiency.} We compare computational cost and inference speed using followings:
        \begin{itemize}[leftmargin=*, topsep=0mm]
            \item \textit{Runtime (ms).} Measures the time required for performing optical flow estimation.
            \item \textit{FLOPs (G).} Represents the total floating-point operations, reflecting computational efficiency.
        \end{itemize}
\end{itemize}

\noindent \textbf{Implementation Details.}
The experiments are carried out using PyTorch version 2.0.0 and Python version 3.8, and the hardware setup consisted of a single Nvidia RTX 3090 GPU with 24GB of memory, alongside a 14-core Intel(R) Xeon(R) Platinum 8362 CPU running at 2.80GHz. More specifically, we pre-train the variational autoencoder (VAE) to obtain the optimal text embedding, and we also use the pre-trained stable diffusion model without modifying its parameters. 

\subsection{Quantitative Results}

We evaluate our method in three key areas: accuracy, diversity, and efficiency. Accuracy is measured using End-Point Error (EPE), Angular Error (AE), and outlier rates (F1-bg, F1-fg, F1-all) for a fair comparison with baselines. Since our approach generates probabilistic flow distributions, we assess prediction diversity using entropy, which quantifies the range of plausible motions from a single image. Finally, we analyze computational efficiency by comparing FLOPs and runtime across datasets. The following sections provide detailed analyses, demonstrating our method’s strong accuracy, diverse predictions, and lower computational cost.

\noindent \textbf{Quantitative Evaluation of Optical Flow Accuracy.}
\FloatBarrier
\begin{table*}[t]
    \caption{Quantitative comparison of optical flow methods on Sintel (Clean/Final), KITTI 2015, and Spring datasets \cite{black2024, Geiger2012CVPR, spring2023}. We evaluate each method using End-Point Error (EPE), Angular Error (AE), and F1-all (Percentage of Outliers), where lower values indicate better performance. F1-bg and F1-fg represent the background and foreground outlier rates, respectively. We highlight the best result in bold and the second-best result with an underline. Additionally, for our method (\ProbDiffFlow), we report the relative improvement over the second-best method in the last row.}
    \label{tab:comparison}
    \centering
    \resizebox{\textwidth}{!}{
    \begin{tabular}{lcccccccccc}
        \toprule
        \multirow{2}{*}{Method}  &\multicolumn{5}{c}{Sintel.C} & \multicolumn{5}{c}{Sintel.F}  \\ 
        \cmidrule(lr){2-6} \cmidrule(lr){7-11}
          & EPE(px) $\downarrow$ & F1-bg(\%) $\downarrow$ & F1-fg(\%) $\downarrow$  & F1-all(\%) $\downarrow$  & AE($^\circ$) $\downarrow$  
          & EPE(px) $\downarrow$ & F1-bg(\%) $\downarrow$  & F1-fg(\%) $\downarrow$ & F1-all(\%) $\downarrow$ & AE($^\circ$) $\downarrow$ \\ 
        \midrule
        RAFT \cite{Teed2020} & 7.14 & 12.15 & 24.30 & 21.80  & 81.85 & 8.32 & 13.42 & 26.85 & 24.97 & 95.35 \\
        PWC-Net \cite{Sun2018} & 8.13  & 14.28 & 28.56 & 25.62 & 93.17 & 10.07 & 15.19 & 30.38 & 28.25 & 115.4\\
        RPK-Net \cite{RPKNet2024Morimitsu} & 3.73 & 10.28 & 20.56 & 18.44 & 42.75 & 6.76 & 10.82 & 21.63 & 20.12 & 77.47 \\
        RAPID-Flow \cite{RAPIDFlow2024} & \underline{3.63} & \underline{9.91} & \underline{19.81}  & \underline{17.77} & \underline{41.60} & \underline{6.52}  & \underline{10.74} & \underline{21.48} & \underline{19.98} & \underline{74.72}\\
        U-Net \cite{Unet2015} & 10.99 & 19.35 & 38.71 & 34.72 & 115.95 & 13.58 & 20.0 & 40.0 & 37.20  & 115.63\\
        Motion-Blurred~\cite{Argaw2021} & 4.27 & 11.80 & 23.58 & 20.19 & 42.91 & 7.16 & 12.40 & 24.80 & 20.98 & 77.69 \\ 
        \midrule
        \multirow{2}{*}{\ProbDiffFlow~(Ours)}
         & \textbf{3.43} & \textbf{9.64} & \textbf{19.28} & \textbf{17.29}  & \textbf{39.31} & \textbf{6.18} & \textbf{10.34} & \textbf{20.69} & \textbf{19.24} & \textbf{70.82} \\
         & 5.51\%  & 2.72\% & 2.68\%  & 2.70\%  & 5.50\%  & 5.21\%  & 3.72\%  & 3.68\%  & 3.79\%  & 5.22\% \\ 
        \bottomrule
        \toprule
        \multirow{2}{*}{Method}  &\multicolumn{5}{c}{KITTI 2015} & \multicolumn{5}{c}{Spring}  \\ 
        \cmidrule(lr){2-6} \cmidrule(lr){7-11}
          & EPE(px) $\downarrow$  & F1-bg(\%) $\downarrow$ & F1-fg(\%) $\downarrow$  & F1-all(\%) $\downarrow$  & AE($^\circ$) $\downarrow$ 
          & EPE(px) $\downarrow$  & F1-bg(\%) $\downarrow$ & F1-fg(\%) $\downarrow$  & F1-all(\%) $\downarrow$ & AE($^\circ$) $\downarrow$\\ 
        \midrule
        RAFT \cite{Teed2020}  & 24.04 & 22.11 & 44.22 & 28.44 & 108.92 & 6.29 & 13.76 & 27.52 & 18.92 &  72.08\\
        PWC-Net \cite{Sun2018} & 32.76 & 26.2 & 52.4 & 33.70 & 114.43  & 7.44  & 16.51 & 33.02 & 22.70 & 85.26 \\
        RPK-Net \cite{RPKNet2024Morimitsu}  & \underline{16.65}  & \underline{17.94} & \underline{35.81}  & \underline{23.03} & \underline{95.29} & 3.06  & 11.56 & 23.13 & 15.90 & 45.07\\
        RAPID-Flow \cite{RAPIDFlow2024}  & 16.80  & 20.0 & 39.90  & 25.72 & 96.90 & \underline{3.04} & \underline{11.35} & \underline{22.69} & \underline{15.60} & \underline{44.84}\\
        U-Net \cite{Unet2015}  & 43.67 & 39.77 & 79.53 & 51.15 & 126.33 & 9.91 & 21.64 & 43.28 & 29.75 & 113.57\\
        Motion-Blurred~\cite{Argaw2021} & 17.80 & 21.27 & 43.52 & 28.50 & 98.50 & 3.64 & 12.14 & 29.30 & 16.31 & 46.50\\ 
        \midrule
        \multirow{2}{*}{\ProbDiffFlow~(Ours)} 
         & \textbf{15.90} & \textbf{17.90} & \textbf{35.79} & \textbf{23.02} & \textbf{82.87} & \textbf{3.02} & \textbf{11.25} & \textbf{22.5} & \textbf{15.47} & \textbf{44.61}\\
         & 4.50\%  & 0.22\%  & 0.06\%  & 0.04\%  & 13.03\% & 0.66\%  & 0.88\%  & 0.84\%  & 0.83\%  & 0.51\%  \\ 
        \bottomrule
    \end{tabular}
    }
\end{table*}

We quantitatively evaluate the optical flow accuracy of our proposed method (\ProbDiffFlow) against various baseline models across four benchmark datasets: Sintel Clean, Sintel Final, KITTI 2015, and Spring (see~\autoref{tab:comparison}). Our evaluation metrics include End-Point Error (EPE), Angular Error (AE), and outlier percentages (F1-bg, F1-fg, F1-all), where lower values indicate better performance. Additionally, we report the relative improvement of our method over the second-best approach in the last row.

\begin{itemize}[leftmargin=*, topsep=0mm]
    \item \textit{Sintel Clean and Sintel Final.} On the Sintel Clean dataset, \ProbDiffFlow achieves the lowest EPE (3.43 px), outperforming RAFT (7.14 px) and PWC-Net (8.13 px) by 51.9\% and 57.8\%, respectively. Compared to the second-best method, RAPID-Flow (3.63 px), our approach reduces EPE by 5.51\%. Similar improvements are observed in outlier rates: F1-bg (9.64\%, 2.72\% improvement), F1-fg (19.28\%, 2.68\% improvement), and F1-all (17.29\%, 2.70\% improvement). Furthermore, our method achieves the lowest AE (39.31$^\circ$), reducing the angular error by 5.50\% compared to RAPID-Flow. 

    For Sintel Final, which contains more complex motion patterns, all methods exhibit slightly worse performance. Nonetheless, \ProbDiffFlow~remains the best-performing model, achieving the lowest EPE (6.18 px), a 5.21\% improvement over RAPID-Flow (6.52 px). These results demonstrate the robustness of \ProbDiffFlow~in handling diverse optical flow scenarios.
    \item \textit{KITTI 2015.} \ProbDiffFlow~achieves the lowest EPE (15.90 px), outperforming RAFT (24.04 px) and RPK-Net (16.65 px), with a 4.50\% reduction in EPE compared to the latter. Additionally, our method reduces AE to 82.87$^\circ$, representing a 13.03\% improvement over RPK-Net (95.29$^\circ$). In terms of outlier rates, \ProbDiffFlow attains F1-bg (17.90\%) and F1-fg (35.79\%), with relative reductions of 0.22\% and 0.06\%, respectively. The overall outlier rate (F1-all) is also minimized to 23.02\% with a 0.04\% improvement. {\color{blue}While the improvements on \textit{KITTI 2015} across various metrics are limited, this outcome is primarily related to the constrained scene diversity and lack of dynamic content in the dataset. Unlike the \textit{Sintel}, which contains cinematic dynamics with diverse camera motion, object interaction, and rich scene transitions, \textit{KITTI 2015} consists of forward-facing driving scenes with relatively static layout and repetitive structure. These characteristics limit the diversity of plausible future frames and reduce the effectiveness of our generative modeling approach. Despite these constraints, \textit{ProbDiffFlow} still consistently outperforms all baselines, highlighting its robustness.}
    \item \textit{Spring.} \ProbDiffFlow~achieves the lowest EPE (3.02 px), outperforming RAPID-Flow (3.04 px) by 0.66\%. Furthermore, AE is reduced to 44.61$^\circ$ with a 0.51\% improvement. Our method also achieves the best outlier rates: F1-bg (11.25\%), F1-fg (22.50\%), and F1-all (15.47\%), improving upon RAPID-Flow by 0.88\%, 0.84\%, and 0.83\%, respectively. 
    {\color{blue}The relatively small improvement on the \textit{Spring} dataset can be attributed to limitations in the textual prompt diversity and the stylized nature of the dataset itself. Specifically, the textual prompt space for \textit{Spring} is relatively restricted, as most images are constrained to generic prompts such as ``mountain scenery'' or ``a landscape with trees'', since most images in \textit{Spring} are high-resolution, stylized environmental scenes, this lack of prompt diversity limits the expressiveness of the CLIP-based embedding, weakens the semantic guidance during generation and restricts the diverse sampling in the latent space. Despite these limitations, \textit{ProbDiffFlow} still performs well across all evaluation metrics.}
    \item \textit{Overall Performance.} Across all datasets and metrics, \ProbDiffFlow~consistently achieves the best accuracy without requiring task-specific training. Our model demonstrates strong generalization across both synthetic datasets (Sintel, Spring) and real-world driving datasets (KITTI 2015). Compared to RAPID-Flow and RPK-Net, two recent high-performing optical flow methods, \ProbDiffFlow~achieves lower EPE, AE, and F1-related across all scenarios, indicating its robustness in handling diverse motion patterns and challenging optical flow estimation tasks.
\end{itemize}

\begin{table*}[t!]
    \centering
    \caption{Computational efficiency comparison of optical flow methods on Sintel.Clean, Sintel.Final (1024×436), KITTI 2015 (1242×375), and Spring (1920×1080) datasets. FLOPs (G) and Time (ms) are measured on a system with a single V100-32GB GPU and an Intel Xeon Gold 6130 CPU (6 vCPUs). Execution time is reported as `training time + inference time (a+b)', while our method is training-free and thus only has inference time. We highlight the best result in bold and the second-best result with an underline. Additionally, for our method (\ProbDiffFlow), we report the relative reduction in FLOPs and inference time compared to the second-best method in the last row.}
    \label{tab:efficiency}
    \begin{tabular}{lcccccccc}
        \toprule
        \multirow{2}{*}{Method}  & \multicolumn{4}{c}{FLOPs (G) $\downarrow$}  & \multicolumn{4}{c}{Time (ms) $\downarrow$} \\ 
        \cmidrule(lr){2-5} \cmidrule(lr){6-9}
        & Sintel.C & Sintel.F & KITTI 2015 & Spring  & Sintel.C & Sintel.F & KITTI 2015 & Spring \\ 
        \midrule
        RAFT~\cite{Teed2020} & 805  & 824  & 840  & 3,739  & 79.2 + 20.4  & 81.3 + 20.8  & 83.1 + 21.4  & 372.7 + 93.7 \\
        PWC-Net~\cite{Sun2018} & 180  & 185  & 187  & 836  & 18.1 + 4.2  & 18.7 + 4.5  & 19.3 + 5.8  & 83.9 + 21.6 \\
        RPK-Net~\cite{RPKNet2024Morimitsu} & 200  & 205  & 209  & 929  & 20.2 + 5.1  & 20.4 + 5.6  & 21.6 + 5.8  & 92.7 + 23.3 \\
        RAPID-Flow~\cite{RAPIDFlow2024} & \underline{128}  & \underline{130}  & \underline{133}  & \underline{594}  
        & \underline{13.3 + 3.1}  & \underline{14.7 + 3.2}  & \underline{17.4 + 2.9}  & \underline{59.3 + 15.8} \\
        U-Net~\cite{Unet2015} & 300  & 308  & 313  & 1,393  & 30.2 + 7.3  & 30.4 + 7.8  & 31.8 + 8.5  & 138.6 + 35.3 \\
        Motion-blurred~\cite{Argaw2021} & 192  & 195  & 203  & 900  & 22.1 + 4.3  & 22.4 + 4.3  & 20.4 + 5.6  & 92.8 + 24.1 \\
        \midrule
        \multirow{2}{*}{\ProbDiffFlow~(Ours)} & \textbf{80}  & \textbf{81}  & \textbf{83}  & \textbf{372}  
        & \textbf{8.9}  & \textbf{9.4}  & \textbf{10.3}  & \textbf{46.8} \\ 
        & 37.5\%  & 37.7\%  & 37.6\%  & 35.2\%  & 45.7\%  & 47.5\%  & 49.2\%  & 37.7\% \\ 
        \bottomrule
    \end{tabular}
\end{table*}
\noindent \textbf{Computational Efficiency Analysis.}
We further analyze the computational efficiency of \ProbDiffFlow~against some baselines, as summarized in~\autoref{tab:efficiency}. Our approach exhibits a significant advantage by being completely training-free, meaning it only requires inference time, whereas all baseline methods involve both training and inference. This results in substantially lower total computational cost across all datasets.

On the Sintel Clean dataset, \ProbDiffFlow~achieves the lowest total runtime of just 8.9 ms, whereas the second-best method, RAPID-Flow, requires 16.4 ms (13.3 ms training + 3.1 ms inference), leading to a 45.7\% reduction in execution time. A similar trend is observed on KITTI 2015 (10.3 ms vs. 20.3 ms, 49.2\% reduction) and Spring (46.8 ms vs. 75.1 ms, 37.7\% reduction), demonstrating the superior computational efficiency of \ProbDiffFlow.

Additionally, \ProbDiffFlow~maintains the lowest FLOPs consumption, requiring only 80G FLOPs on Sintel Clean, compared to 128G FLOPs for RAPID-Flow, achieving a 37.5\% reduction. This represents a substantial improvement over RAFT, which consumes 805G FLOPs, making our approach nearly 10$\times$ more efficient. On the Spring dataset, \ProbDiffFlow~remains highly efficient at 372G FLOPs, compared to 594G FLOPs for RAPID-Flow, resulting in a 35.2\% reduction. Notably, RAFT's computational demand on Spring reaches an overwhelming 3,739G FLOPs, underscoring the lightweight nature of our approach. Furthermore, on the real-world KITTI 2015 dataset, \ProbDiffFlow~requires just 83G FLOPs, significantly lower than the 133G FLOPs of RAPID-Flow, achieving a 37.6\% reduction. 

In summary, our method not only eliminates the need for costly training but also achieves the fastest total runtime while maintaining competitive accuracy. This makes \ProbDiffFlow~particularly attractive for real-time processing and computationally constrained environments.

\begin{table}[t]
    \centering
    \caption{Entropy evaluation of optical flow diversity across different datasets. A higher entropy value (\(\uparrow\)) indicates greater flow diversity, where 0 represents a fully deterministic prediction and 1 indicates infinitely diverse results. Entropy values are normalized to the range \((0,1]\). We highlight the best result in bold and the second-best result with an underline.}
    \label{tab:entropy}
    \begin{tabular}{lc}
        \toprule
        Dataset & Entropy \(\uparrow\)  \\
        \midrule
        Sintel.C \cite{black2024}    & 0.81 \\
        Sintel.F \cite{black2024}    & \underline{0.84}  \\
        KITTI2015 \cite{Geiger2012CVPR}  & \textbf{0.85} \\
        Spring \cite{spring2023}     & 0.79  \\
        \bottomrule
    \end{tabular}
\end{table}

\noindent \textbf{Diversity Analysis.}
We further evaluate the capability of our model to generate diverse optical flow predictions using entropy as an indicator. As shown in~\autoref{tab:entropy}, our method achieves high entropy scores across all datasets, confirming its effectiveness in capturing multiple plausible motion outcomes. Notably, our model attains the highest entropy value of 0.85 on the real-world KITTI 2015 dataset, highlighting its strength in generating diverse predictions in realistic driving environments. The Sintel Final dataset closely follows (with 0.84 entropy), demonstrating robust performance in complex synthetic conditions. Overall, these entropy values reflect that our model effectively addresses inherent motion uncertainties, producing a rich variety of plausible and physically consistent optical flow estimations rather than deterministic single outcomes.

\subsection{Qualitative Results}
\begin{figure*}[t!]
    \centering
  \includegraphics[width=1.0\linewidth]{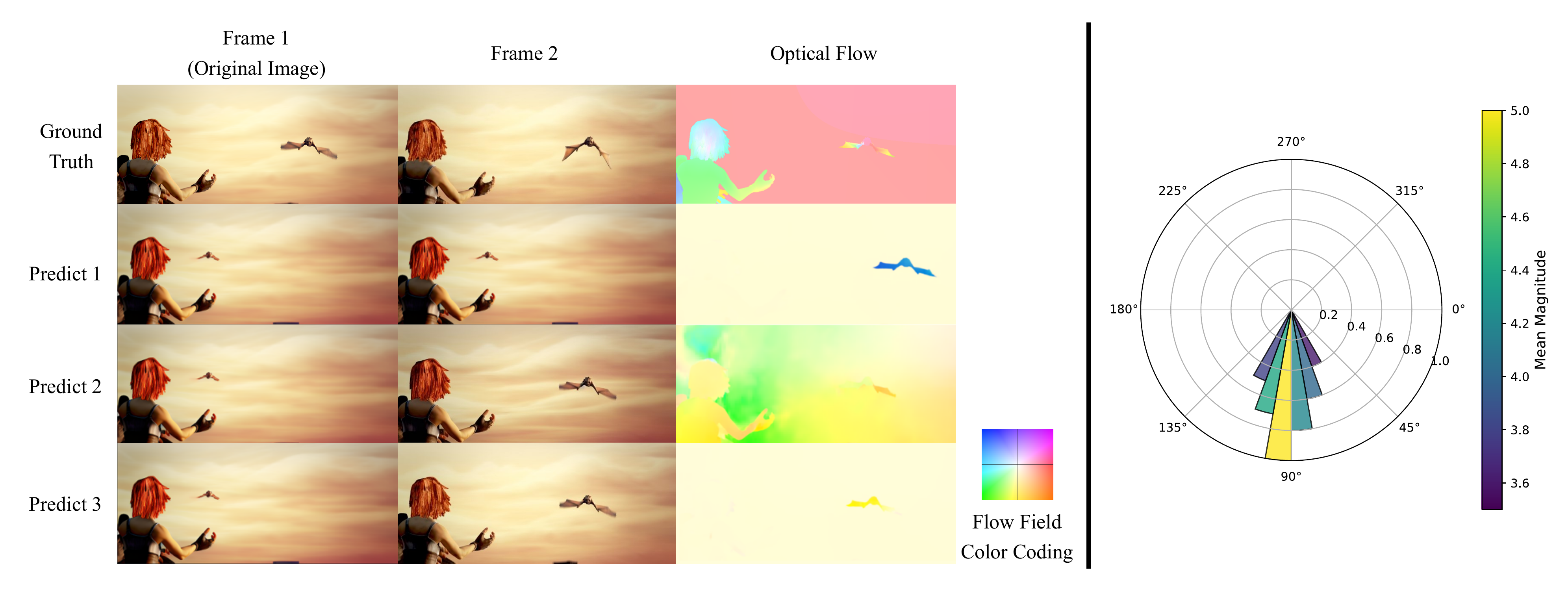}
  \caption{\textbf{Left:} Optical flow visualization for a single image. The first column shows the original image (Frame 1). The first row presents the ground truth for Frame 2 and the corresponding optical flow, while the second to fourth rows show different predictions generated by our method, illustrating the variations in both Frame 2 and the predicted optical flow fields. The flow field color coding follows a standard convention \cite{baker2007}. \textbf{Right:} Optical flow direction and magnitude Analysis for a single image. This visualization shows the distribution of predicted optical flow directions and their respective magnitudes for each pixel in the image. The histogram’s length represents the number of pixels moving in that direction, while the color indicates the mean flow magnitude.}
  \label{fig:single_image_analysis}
\end{figure*}

We further evaluate our method qualitatively through single-image analysis and comparisons with baseline methods. The single-image analysis visualizes the predicted motion distributions and flow variations from a single input, highlighting the diversity captured by our approach. In contrast, the method comparison section presents visualization results on both synthetic (Sintel) and real-world (KITTI 2015) datasets, demonstrating how our method performs against two-frame and single-frame baselines.

\noindent \textbf{Single Image Analysis.} The goal of this experiment is to visualize and analyze the different predicted flow results from the same input image, highlighting the various plausible outcomes generated by our approach. As presented in~\autoref{fig:single_image_analysis}, the right part shows the results, where the first row corresponds to the ground truth. In the ground truth optical flow, we observe that the bird's wings are flapping downward, the girl is extending her arms, and the background shows a general rightward movement, indicating that the camera is panning to the right. In the generated predictions (rows 2-4), the following trends can be observed:
\begin{itemize}[leftmargin=*, topsep=0mm]
    \item Predict 1: The bird tends to fly to the left, while the background shows a slight downward motion, suggesting that the camera is moving slightly upward.
    \item Predict 2: The flow is more complex, the girl shows a downward motion, while the left part of the background tends to move toward the lower-left corner.
    \item Predict 3: The bird moves downward, and the background also shows a slight downward movement, indicating upward camera motion.
\end{itemize}
The background movement across all predictions remains largely consistent, reflecting a stable flow pattern associated with the camera movement. In contrast, the movement of the foreground objects varies between predictions, showing the diverse possibilities generated by our method. These visual results demonstrate our model's ability to capture multiple plausible outcomes and illustrate the inherent uncertainty and multiple possible future states for dynamic scenes.

As observed from the left image in~\autoref{fig:single_image_analysis}, most of the predicted optical flows exhibit a downward direction with relatively large mean magnitudes, as indicated by the yellow region. Other directions show less significant movement and smaller magnitudes. This suggests that the predicted motion is primarily downward, reflecting the consistent movement of key objects in the scene.

\begin{figure*}[t!]
    \centering
  \includegraphics[width=1.0\linewidth]{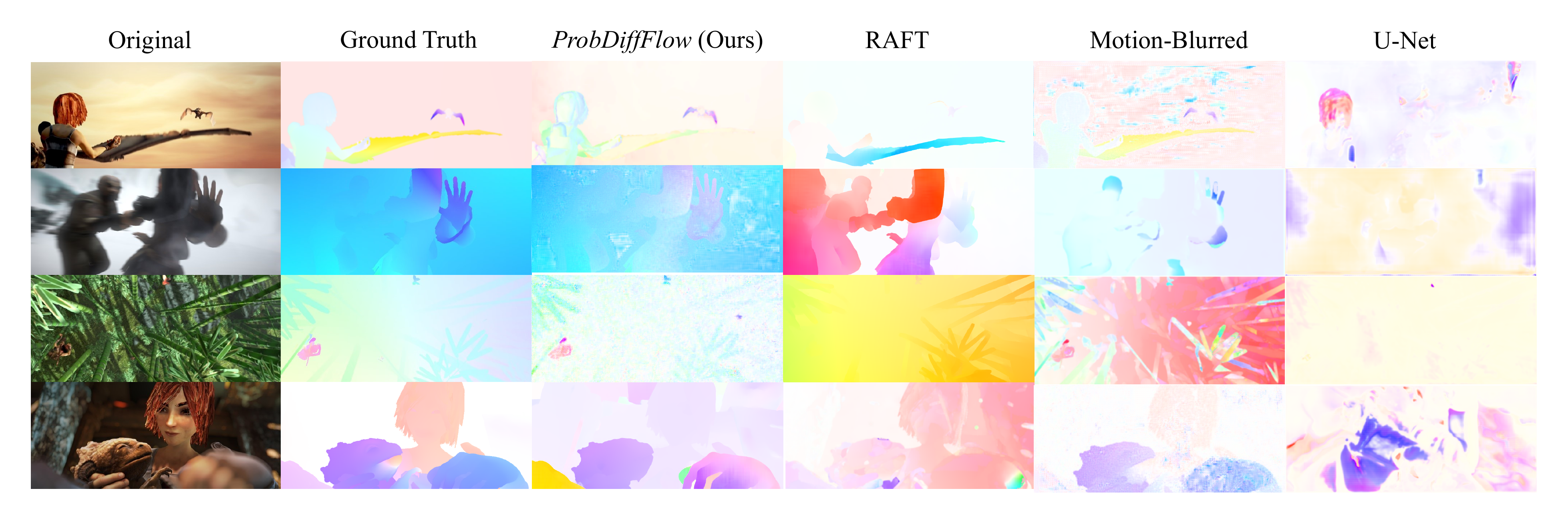}
  \caption{Qualitative comparison of optical flow estimation on the MPI Sintel dataset \cite{black2024}. From left to right: (a) original image, (b) ground truth flow, (c) \ProbDiffFlow~(Ours), (d) RAFT~\cite{Teed2020}, (e) Motion-Blurrd~\cite{Argaw2021}, (f) U-Net~\cite{Unet2015}. The flow color encodes motion direction and magnitude.}
  \label{fig:methods_compare_sintel}
\end{figure*}

\begin{figure*}[t!]
    \centering
  \includegraphics[width=1.0\linewidth]{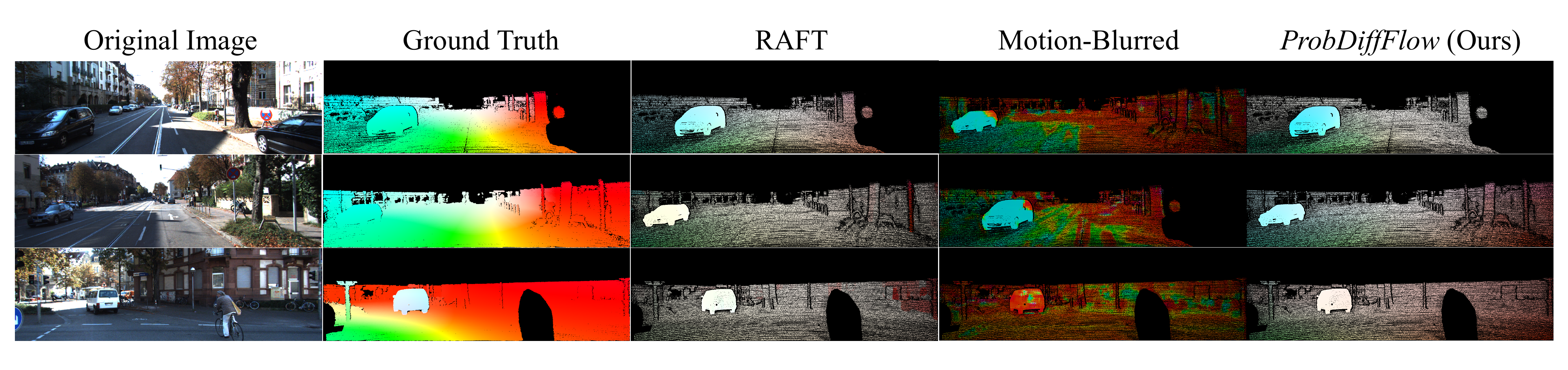}
  \caption{Qualitative Comparison on the KITTI 2015 Dataset (Including Occlusions) \cite{Geiger2012CVPR}. We show three samples from the KITTI 2015 dataset. From left to right: (a) the original input image, (b) the ground truth flow with occlusions (in black regions), (c) predictions by RAFT~\cite{Teed2020}, (d) predictions by Motion-Blurred~\cite{Argaw2021}, (e) predictions by \ProbDiffFlow~(Ours). Areas in black indicate occluded regions where flow is not measured.}
  \label{fig:methods_compare_kitti}
\end{figure*}

\noindent \textbf{Comparison with Different Methods.}
As presented in~\autoref{fig:methods_compare_sintel} and~\autoref{fig:methods_compare_kitti}, a qualitative comparison is shown for different optical flow estimation methods, including our approach, RAFT (two-frame), U-Net (single-image), and Motion-Blurred (single-image). Since our method generates a distribution of flow predictions ($\hat{\Gamma}$), we select the best flow at each pixel based on the EPE metric and visualize.

On the Sintel dataset, a synthetic movie-based benchmark, our method produces results that closely align with the ground truth while preserving sharp motion boundaries and fine details. RAFT, as a two-frame method, generally provides clear boundaries and detailed motion estimations, but our approach achieves comparable or even superior results without requiring multiple frames. In contrast, single-image methods like U-Net and Motion-Blurred struggle to recover motion accurately. U-Net produces highly noisy and unrealistic predictions, making the motion estimation difficult to interpret. As seen in the last column of~\autoref{fig:methods_compare_sintel}, the results are almost unrecognizable, failing to capture any meaningful motion structures. Motion-Blurred, although designed for motion estimation from a single image, produces highly diffused and less precise predictions, especially in regions with fast-moving objects or detailed textures.

On the KITTI 2015 dataset, which consists of real-world driving scenes, our method continues to demonstrate strong performance. Motion-Blurred, a single-frame method, performs notably worse on KITTI 2015, failing to generate reliable predictions in real-world scenarios with complex road environments and occlusions. From the visual results, it struggles to distinguish between roads and sidewalks, as well as between vehicles and their surroundings. Instead, it incorrectly interprets shadows as road boundaries, making its motion predictions highly unreliable. As shown in~\autoref{fig:methods_compare_kitti}, this limitation becomes particularly evident as its predictions become inconsistent. RAFT, on the other hand, benefits from having two frames, leading to well-structured motion predictions with clearer object boundaries and finer details. However, our single-frame approach is capable of producing similarly structured and precise motion estimates, even in occluded and highly dynamic regions.

While quantitative metrics highlight the ability of \ProbDiffFlow~to generate diverse and accurate flow distributions, the qualitative advantage is evident in the visual clarity and precision of individual predictions. Our method outperforms both single-frame and two-frame baselines, achieving better motion estimation in both synthetic and real-world datasets.

\begin{table*}[t]
    \centering
    \caption{Ablation study on the effects of text inversion, sample distance, and sample number on optical flow performance. EPE represents the End-Point Error, while AE refers to the Angular Error (measured in degrees \(^\circ\)). Entropy quantifies flow diversity, with higher values indicating broader distributions, normalized to \((0, 1]\). Bold denotes the best result, and underline represents the second-best result for each metric.}
    \label{tab:abaltion study}
    \begin{tabular}{cccccc}
        \toprule
        Text Inv. & Sample Distance & Sample Number & EPE (Mean) & AE(Mean)  & Entropy \\
        \midrule
        \ding{51} & 30 & 500 & 8.18 & \underline{80.23$^\circ$} & 0.81 \\
        \ding{55} & 30 & 500 & 8.55 & 85.10$^\circ$ & \underline{0.87} \\
        \ding{51} & 30 & 1000 & \textbf{8.10} & \textbf{79.50$^\circ$} & 0.78 \\
        \ding{55} & 30 & 1000 & 8.48 & 84.30$^\circ$ & 0.85 \\
        \ding{51} & 50 & 500 & 8.30 & 82.00$^\circ$ & 0.83 \\
        \ding{55} & 50 & 500 & 8.65 & 86.40$^\circ$ & \textbf{0.89} \\
        \ding{51} & 50 & 1000 & \underline{8.20} & 81.30$^\circ$ & 0.80 \\
        \ding{55} & 50 & 1000 & 8.60 & 85.50$^\circ$ & 0.86 \\
        \bottomrule
    \end{tabular}
\end{table*}

\subsection{Ablation Study}
We conduct an ablation study on the Sintel Clean dataset to evaluate the impact of different factors on optical flow estimation performance. Specifically, the study analyzes the effect of text inversion, sampling distance, and sample number on key metrics such as EPE, Angular Error, and Entropy.

Text inversion refers to the optimization step used to refine the latent text embeddings. Sample distance represents the defined distance in the latent space used during the nearby sampling process, while sample number refers to the number of generated nearby samples. From the results in~\autoref{tab:abaltion study}, we identify the following key findings:
\begin{itemize}[leftmargin=*, topsep=0mm]
    \item \textit{Text Inversion.} Applying text inversion consistently improves performance across all metrics. For example, the lowest EPE (8.10) and angular error (79.50$^\circ$) occur when text inversion is applied, demonstrating that optimized text embeddings lead to more accurate flow predictions. Without text inversion, both EPE and AE increase significantly, reducing the precision of optical flow estimation.
    \item \textit{Sample Distance.} A larger sample distance (50) increases both EPE and angular error, reducing accuracy in flow estimation. However, it results in higher entropy, indicating a greater diversity of optical flow predictions. In contrast, a shorter sample distance (30) leads to lower entropy but improves accuracy by reducing both EPE and AE.
    \item \textit{Sample Number.} Increasing the sample number from 500 to 1000 slightly reduces EPE, indicating that a larger number of nearby sampled latent states generally captures more accurate flows. However, entropy gradually decreases as the sample number increases, implying a trade-off between motion diversity and prediction precision.
\end{itemize}

Overall, the ablation study indicates that incorporating text inversion, selecting an optimal sample distance, and increasing sample numbers significantly influence the accuracy, diversity of the estimated optical flow, and the ability to capture motion dynamics.

{\color{blue}\noindent\textbf{Parameter Selection Summary.} To ensure reproducibility and practical applicability, we briefly explain the rationale behind the chosen parameters. The sample distances (30 and 50) are selected based on empirical trials, representing a reasonable trade-off between perturbation strength and latent coherence in the spherical latent space. Smaller distances ($<30$) led to insufficient diversity, while larger values ($>50$) resulted in perceptual artifacts in generated images. The sample numbers (500 and 1000) are chosen to reflect computational feasibility under real-time constraints and to observe the marginal utility of more samples. We found that increasing sample numbers beyond 1000 yields diminishing returns while increasing inference cost linearly. As such, these parameters provide a robust setting for balancing diversity and precision in probabilistic flow estimation.}

\section{Related Work}

\noindent \textbf{Optical Flow.}
Optical flow estimation methods can be broadly categorized into two paradigms: two-frame and single-image approaches. Two-frame methods predict dense motion fields between consecutive frames, leveraging rich temporal information, while single-image methods estimate optical flow from a single frame, relying on learned motion priors and underlying scene structures.

Traditional two-frame methods formulate optical flow estimation as an optimization problem involving energy minimization. These methods balance a data fidelity term with a regularization term within a variational framework to infer a dense flow field~\cite{Horn1981}. Feature-based methods utilize a pyramid approach to match features across frames efficiently. Discrete optimization techniques construct cost volumes and apply global optimization strategies to generate dense flow fields~\cite{Brox2011, Chen2016, menze2015discrete, Xu2017}. More recently, deep learning has been incorporated into optical flow estimation. FlowNet introduces CNN to predict flow directly from image. pairs
Subsequent models such as SPyNet and PWC-Net refine accuracy by employing coarse-to-fine pyramidal processing.
RAFT further improves performance by constructing multi-scale 4D correlation volumes and iteratively updating a single high-resolution flow field~\cite{Teed2020}. Recent advances also explore Transformer-based architectures for optical flow estimation~\cite{Kumar2024Survey}. CRAFT integrates Transformers into a cost aggregation framework to improve global context modeling~\cite{sui2022craftcrossattentionalflowtransformer}. GMFlow leverages global matching to align features across frames.
FlowFormer introduces a token propagation module for iterative refinement~\cite{huang2022flowformertransformerarchitectureoptical}, and TransFlow combines self- and cross-attention to enhance motion reasoning~\cite{lu2023transflowtransformerflowlearner}. More recent models focus on enhancing efficiency and generalization. RPKNet integrates partial kernel convolutional layers and separates large kernels to achieve high performance with reduced computational costs~\cite{RPKNet2024Morimitsu}. RAPIDFlow introduces NeXt1D convolution blocks to maintain high-quality flow estimation while improving generalization and efficiency~\cite{RAPIDFlow2024}. 

While most optical flow estimation methods rely on two consecutive frames, some approaches estimate dense flow fields from a single image. One notable method employs a convolutional neural network (CNN) to predict future motion directly from a static image by learning motion priors from large-scale video datasets~\cite{2015singleflowwalker}. This approach does not require human labeling and generalizes well across diverse scenarios by capturing motion context from still images. Another approach utilizes an encoder-decoder network to infer motion from a motion-blurred single image~\cite{Argaw2021}. DepthStillation synthesizes optical flow from a single still image by leveraging a monocular depth estimation network to generate new views~\cite{Aleotti2021}. {\color{blue}However, these single-image methods typically require supervised training and produce deterministic predictions based on learned priors. In contrast, our approach is training-free and probabilistic. Instead of learning a deterministic motion map, we generate a distribution of plausible motions from a single image by leveraging generative modeling, eliminating the need for large labeled datasets.}

\noindent \textbf{Diffusion Model.}
In recent years, diffusion models have emerged as a powerful approach for generative modeling~\cite{croitoru2023diffusion}. These models operate by iteratively corrupting an image with noise and then learning to reverse the process to generate high-quality samples. Recent studies have also explored diffusion models in structured prediction tasks such as object detection~\cite{chen2023diffusiondet}.

Denoising diffusion probabilistic models (DDPM) represent a class of latent variable models that gradually add Gaussian noise to an image and learn to reverse this process to produce high-quality outputs~\cite{ho2020denoising}. Denoising diffusion implicit models (DDIM) redefine the reverse process by breaking the continuous Markov chain of DDPM, enabling more efficient sampling and accelerating inference~\cite{song2022denoising}. The stable diffusion model extends this paradigm by introducing a latent diffusion framework, which compresses images into a latent space before applying the diffusion process. This approach significantly enhances computational efficiency while maintaining high-quality generation, making it widely used in various vision applications~\cite{stablediffusion}.

{\color{blue}Existing diffusion-based applications primarily focus on image generation, editing, or detection tasks. In contrast, we adapt diffusion models to the task of optical flow prediction by combining latent-space nearby sampling with motion estimation. Our framework utilizes diffusion not to synthesize final images, but to explore plausible future states in latent space, from which probabilistic motion distributions are inferred without any supervised training.}

\section{Conclusion}
In this work, we propose \ProbDiffFlow, an efficient training-free probabilistic optical flow estimation framework that generates motion distributions from a single image. Unlike existing single-frame approaches that rely on supervised labeled training and produce deterministic predictions, our method leverages a diffusion-based generative model to synthesize plausible future frames and estimate optical flow distributions without requiring labeled datasets. This formulation enables our model to capture multiple plausible motions while maintaining high accuracy and computational efficiency. Extensive experiments on both synthetic and real-world datasets highlight the superior performance of our method.

Despite its advantages, \ProbDiffFlow~has certain limitations. First, while our method predicts a distribution of possible motions, it does not model the physical forces behind motion, which may limit its interpretability in physics-based applications. {\color{blue}Second, since the framework relies on a pre-trained optical flow model (RAFT) for estimation, its performance may be influenced by domain biases and sensitivity to generation artifacts. Specifically, RAFT was trained on datasets such as FlyingChairs, Sintel, and KITTI, covering common scenarios like urban and indoor scenes. When applied to out-of-domain distributions, such as medical or aerial imagery, the pre-trained model may exhibit degraded performance. Furthermore, diffusion-generated frames may contain artifacts such as blurred boundaries or texture distortions, which could affect the accuracy and stability of flow estimation.}

To address these limitations, future work could incorporate physics-based constraints to guide motion predictions towards more realistic behaviors and explore strategies to adapt the framework to a wider range of real-world domains.

\begin{acknowledgement}
Kai Wang is the corresponding author and is supported by NSFC 62302294 and NSFC U2241211. Xuemin Lin is supported by NSFC U20B2046 and NSFC U2241211. Wenjie Zhang is supported by ARC DP230101445 and ARC FT210100303.
\end{acknowledgement}




\bibliographystyle{fcs}
\bibliography{ref}

\begin{biography}{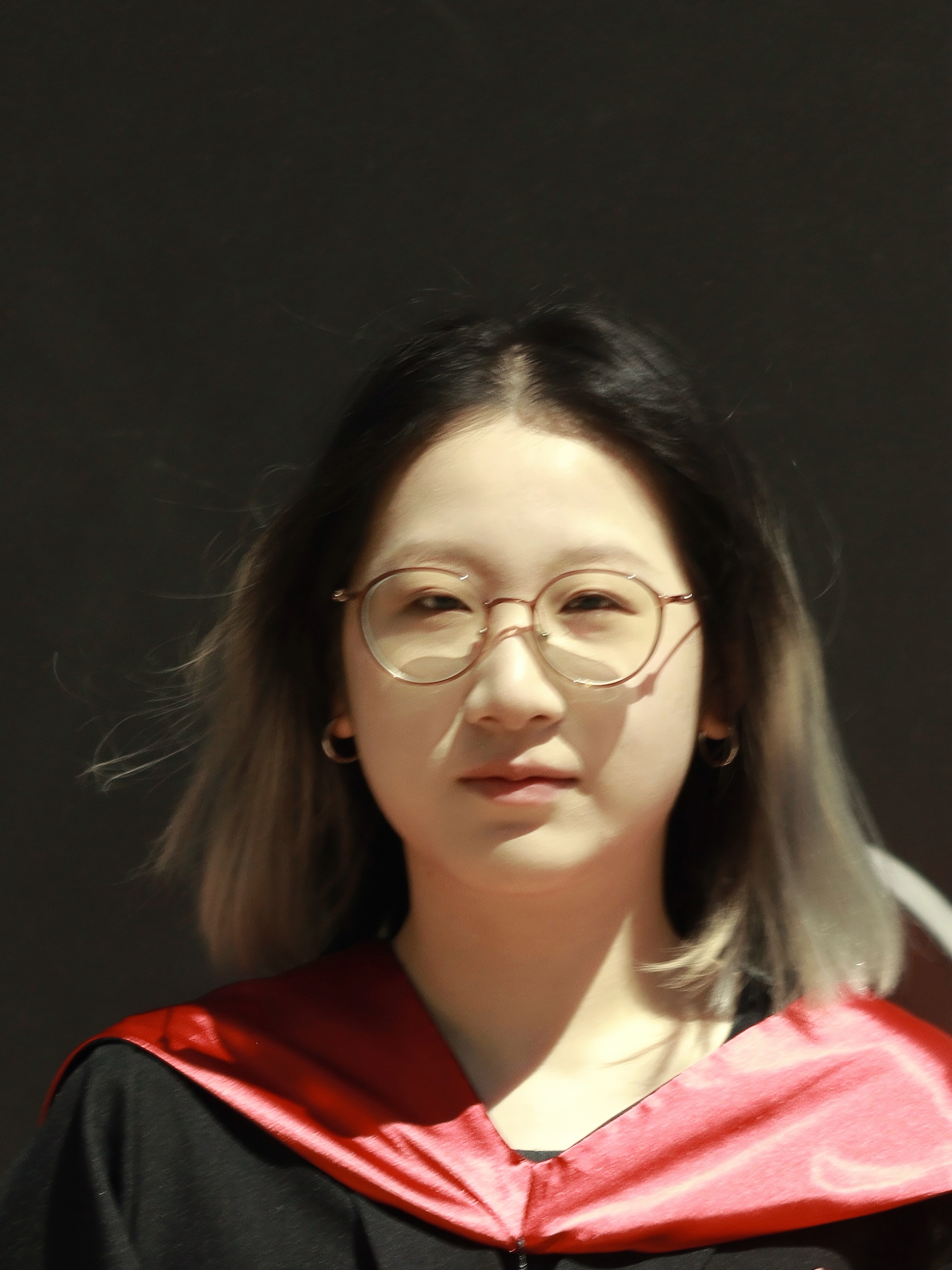}
{Mo Zhou}
received the Bachelor of Advanced Computing (Honours) and Master of Machine Learning and Computer Vision from the Australian National University, Australia.
Her research interests include noise data repair and semantic segmentation.
\end{biography}

\begin{biography}{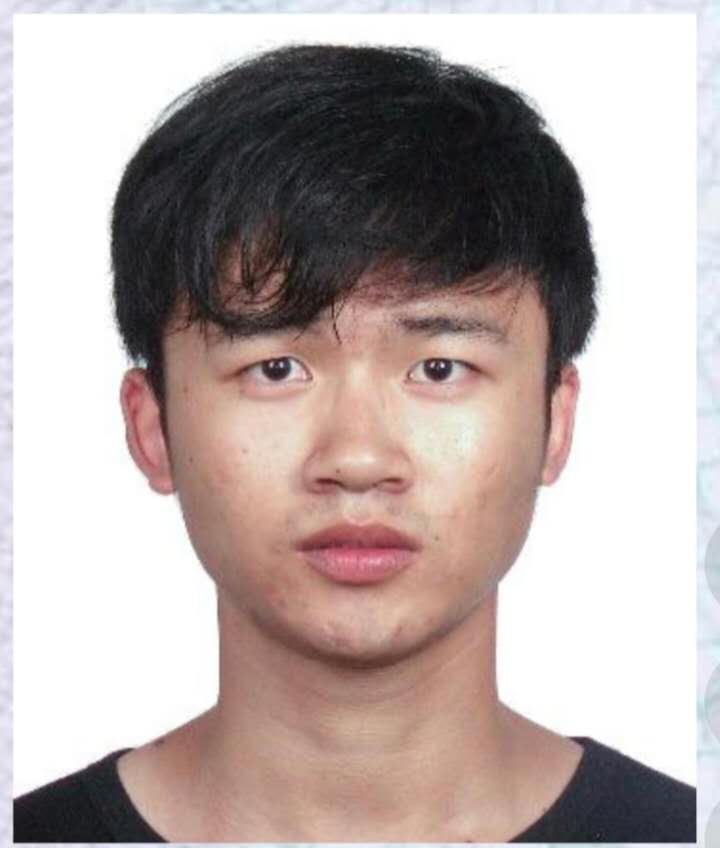}
{Jianwei Wang} 
is currently a PhD candidate in the School of Computer Science and Engineering, University of New South Wales. His research is focused on data quality, graph analytics, and scalable algorithms for dense subgraph discovery.
\end{biography}

\begin{biography}{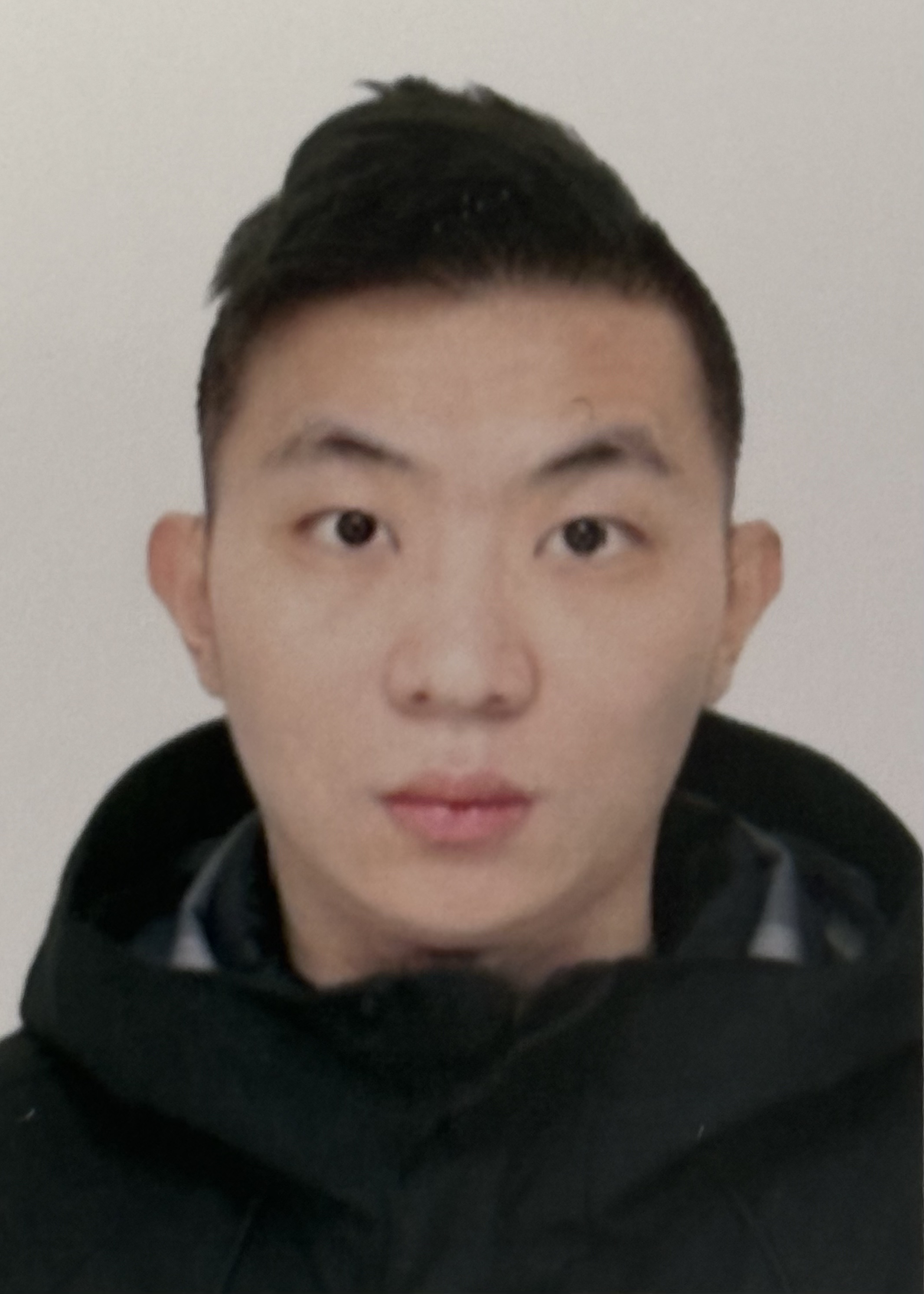}
{Xuanmeng Zhang}
received the B.S. degree from Zhejiang University, China, in 2020. 
His research interests include generative model, image retrieval, depth completion, and 3D generation. He published several papers in CVPR, ICCV and IJCV.
\end{biography}

\begin{biography}{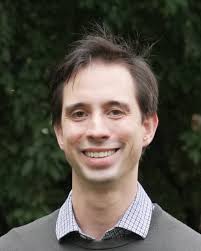}
{Dylan Campbell}
is currently a Lecturer in Computing at the Australian National University. He was previously a Research Fellow at the University of Oxford and the Australian Centre for Robotic Vision. His research interests include 3D vision and geometric deep learning.
\end{biography}
\vspace{0.5em}

\begin{biography}{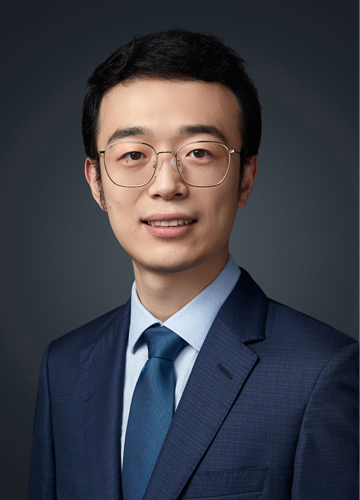}
{Kai Wang} 
is currently an associate professor at Antai College of Economics and Management, Shanghai Jiao Tong University. He received the PhD degree in Computer Science from the University of New South Wales in 2020. His research interests lie in big data analytics, especially for the graph, social network and spatial data.  
\end{biography}
\vspace{-2.5em}

\begin{biography}{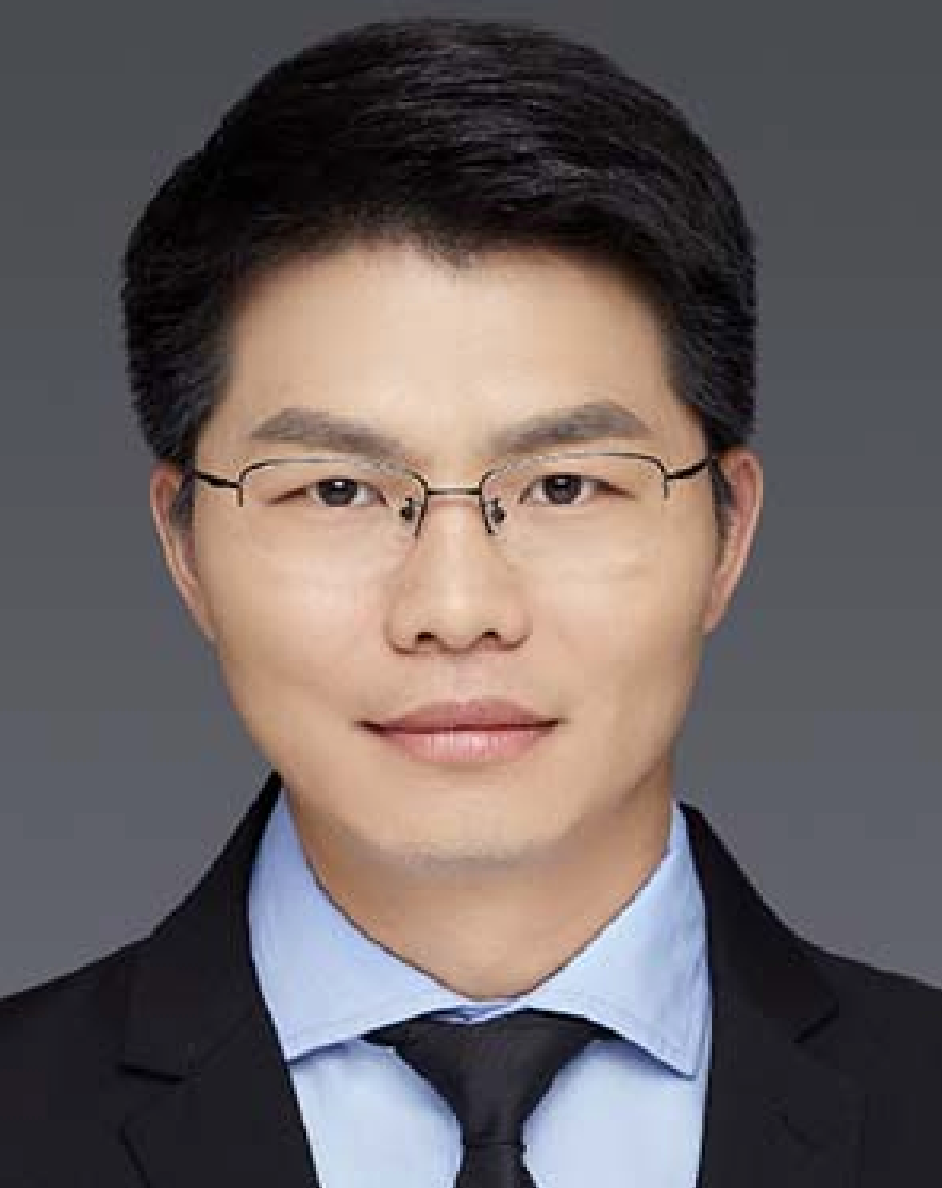}
{Long Yuan} 
received the BS and MS degrees from Sichuan University and the PhD degree from the University of New South Wales. He is currently a professor with the School of Computer Science and Engineering, Nanjing University of Science and Technology. His research interests include graph data management and analysis. 
\end{biography}
\vspace{1.0em}

\begin{biography}{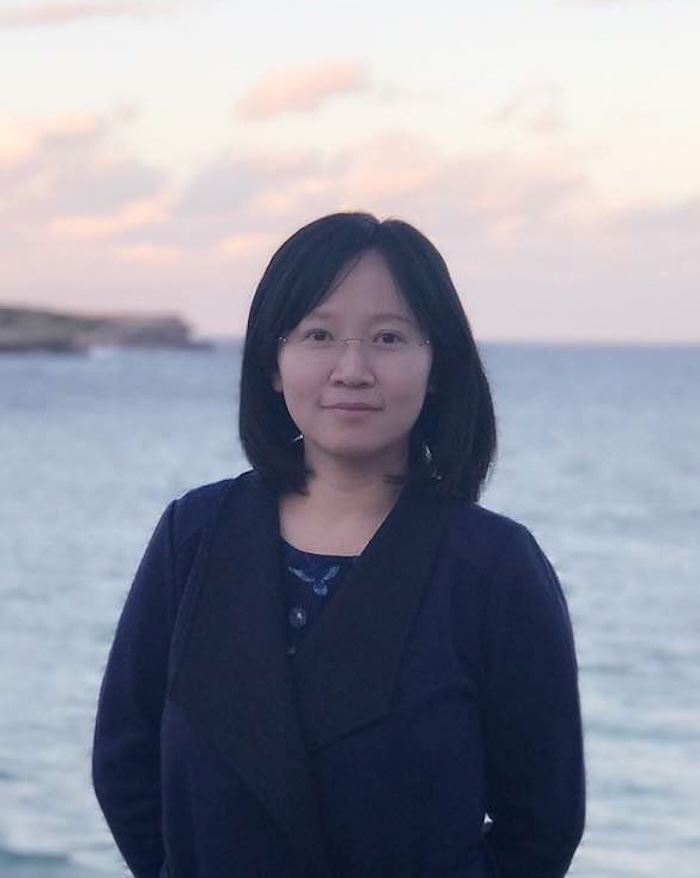}
{Wenjie Zhang} 
received the PhD degree in computer science and engineering from the University of New South Wales in 2010. She is currently a professor and ARC Future Fellow with the School of Computer Science and Engineering, the University of New South Wales, Australia. 
\end{biography}
\vspace{-0.5em}

\begin{biography}{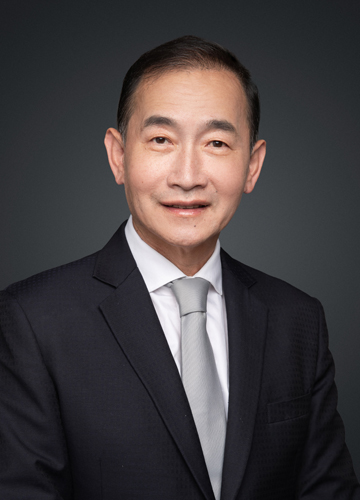}
{Xuemin Lin}  
is a Chair Professor at Antai College of Economics and Management, Shanghai Jiao Tong University. He is Fellow of the IEEE, Member of Academia Europaea and Fellow of Asia-Pacific Artificial Intelligence Association. His main research areas are databases and graph analytics.
\end{biography}

\end{sloppypar}
\end{document}